\documentclass{article}

\usepackage{arxiv}

\usepackage[utf8]{inputenc} 
\usepackage[T1]{fontenc}    
\usepackage{hyperref}       
\usepackage{url}            
\usepackage{booktabs}       
\usepackage{mathptmx} 
\usepackage{amsfonts}       
\usepackage{amsmath} 
\usepackage{amssymb}  
\usepackage{nicefrac}       
\usepackage{microtype}      
\usepackage{cleveref}       
\usepackage{lipsum}         
\usepackage{graphicx} 
\graphicspath{{Figures/}}
\DeclareGraphicsExtensions{.pdf,.jpeg,.png,.eps}
\usepackage{doi}
\usepackage{subfig}
\usepackage{subfloat}
\usepackage{xcolor}
\usepackage{multirow}

\title{Floor extraction and door detection for visually impaired guidance}

\date{}

\author{ \href{https://orcid.org/0000-0003-2674-4844}{\includegraphics[scale=0.06]{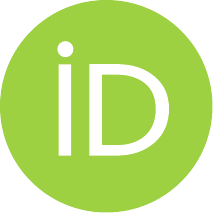}\hspace{1mm}Bruno Berenguel-Baeta}\thanks{Corresponding author.} \\
	Instituto de Investigacion en Ingenieria de Aragon\\
	Department of Computer Science and Systems Engineering\\
	University of Zaragoza,
	Zaragoza, Spain \\
	\texttt{berenguel@unizar.es} \\
	\And
	Manuel Guerrero-Viu \\
	Instituto de Investigacion en Ingenieria de Aragon\\
	Department of Computer Science and Systems Engineering\\
	University of Zaragoza,
	Zaragoza, Spain \\
	\texttt{bermudez@unizar.es} \\
	\And
	Alejandro de Nova \\
	Instituto de Investigacion en Ingenieria de Aragon\\
	Department of Computer Science and Systems Engineering\\
	University of Zaragoza,
	Zaragoza, Spain \\
	\And
	\href{https://orcid.org/0000-0002-8479-1748}{\includegraphics[scale=0.06]{orcid.pdf}\hspace{1mm}Jesus Bermudez-Cameo} \\
	Instituto de Investigacion en Ingenieria de Aragon\\
	Department of Computer Science and Systems Engineering\\
	University of Zaragoza,
	Zaragoza, Spain \\
	\texttt{bermudez@unizar.es} \\
	\And
	\href{https://orcid.org/0000-0002-8949-2632}{\includegraphics[scale=0.06]{orcid.pdf}\hspace{1mm}Alejandro Perez-Yus} \\
	Instituto de Investigacion en Ingenieria de Aragon\\
	Department of Computer Science and Systems Engineering\\
	University of Zaragoza,
	Zaragoza, Spain \\
	\texttt{alperez@unizar.es} \\
	\And
	\href{https://orcid.org/0000-0001-5209-2267}{\includegraphics[scale=0.06]{orcid.pdf}\hspace{1mm}Jose J. Guerrero} \\
	Instituto de Investigacion en Ingenieria de Aragon\\
	Department of Computer Science and Systems Engineering\\
	University of Zaragoza,
	Zaragoza, Spain \\
	\texttt{josechu.guerrero@unizar.es} \\
}
\renewcommand{\shorttitle}{\textit{Floor extraction and door detection}}
\newcommand\blfootnote[1]{%
  \begingroup
  \renewcommand\thefootnote{}\footnote{#1}%
  \addtocounter{footnote}{-1}%
  \endgroup
}
\hypersetup{
pdftitle={Floor extraction and door detection for visually impaired guidance},
pdfsubject={cs.CV},
pdfauthor={B. Berenguel-Baeta, M. Guerrero-Viu, A. Nova, J. Bermudez-Cameo, A. Perez-Yus and J.J. Guerrero},
pdfkeywords={Computer vision; omnidirectional camera; image processing; floor extraction; door detector; RGBD camera; fish-eye camera; visually impaired guidance},
}

\begin{document}
\maketitle
\blfootnote{A final version of this article can be found at \url{https://doi.org/10.1109/ICARCV50220.2020.9305464}}

\begin{abstract}
Finding obstacle-free paths in unknown environments is a big navigation issue for visually impaired people and autonomous robots. Previous works focus on obstacle avoidance, however they do not have a general view of the environment they are moving in. New devices based on computer vision systems can help impaired people to overcome the difficulties of navigating in unknown environments in safe conditions. In this work it is proposed a combination of sensors and algorithms that can lead to the building of a navigation system for visually impaired people.
Based on traditional systems that use RGB-D cameras for obstacle avoidance, it is included and combined the information of a fish-eye camera, which will give a better understanding of the user's surroundings. The combination gives robustness and reliability to the system as well as a wide field of view that allows to obtain many information from the environment.
This combination of sensors is inspired by human vision where the center of the retina (fovea) provides more accurate information than the periphery, where humans have a wider field of view. The proposed system is mounted on a wearable device that provides the obstacle-free zones of the scene, allowing the planning of trajectories for people guidance.

\end{abstract}

\keywords{Computer vision \and omnidirectional camera \and image processing \and floor extraction \and door detector \and RGBD camera \and fish-eye camera \and visually impaired guidance}

\section{Introduction}
\label{sec:intro}

Sight is the sense that gives us more information about our surroundings and the obstacles placed in it. According to \cite{plikynas2020indoor} and the World Health Organization, in 2020 there were 441.5 million people with a visual handicap where 36 million were totally blind. Even though there are different devices that visually impaired people can use to overcome their daily routine, the increase of capacity and portability of vision systems can improve their mobility and independence. These advances of the technology and computer vision systems make even more important the creation of new helping devices. In this work it is proposed a new system for helping visually impaired people to navigate safely through unknown environments.

The related work focus on avoiding obstacles close to the user. These systems allow the user to move freely but it is difficult to plan trajectories in an environment or across different environments.
However, with an appropriate selection of sensors, finding obstacle-free zones and doors to go from one environment to another can allow the planning and creation of safe routes in the environment.

The development of guidance systems is a recurrent research topic since it embraces many possibilities of sensors and algorithms, considering from visually impaired people to autonomous robots. Some guiding systems use devices placed in the environment and others over the user \cite{simoes2016blind,pereira2015blind,dao2016indoor}. In \cite{islam2018indoor}, infrared transmitters are placed in the environment, continuously monitoring the user's position. Even though this approach has a good performance, it is limited to indoor environments. Another system that uses complex sensors is described in \cite{guimaraes2013analysis}, where ultrasonic sensors and a GPS system is attached to a white cane to obtain the position of the user. The main disadvantage is that GPS sensors do not work properly in all the environments, such as indoor environments, where only the ultrasonic sensor would provide information. Besides, ultrasonic sensors provide noisy information, losing accuracy in the detection of obstacles and limiting the movement and the safety of the user. 

\begin{figure}[!t]
    \centering
    \subfloat[ \label{fig:rgb-fish}]{\includegraphics[width=0.24\textwidth]{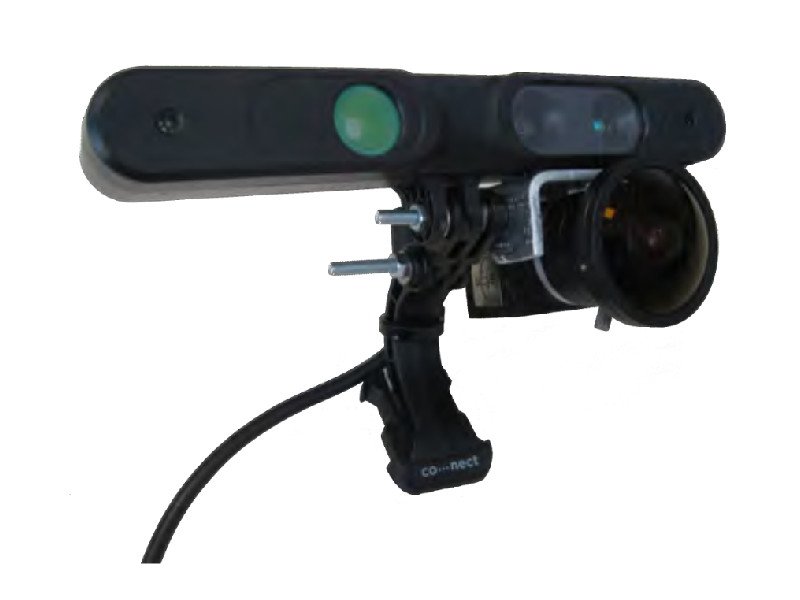}}
    \hfill
    \subfloat[ \label{fig:rgb-fov}]{\includegraphics[width=0.24\textwidth]{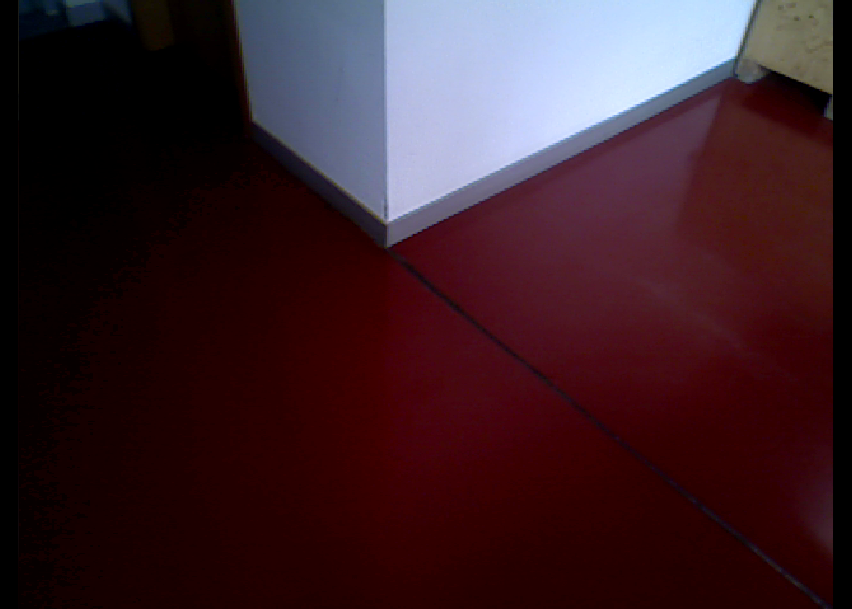}}
    \hfill
    \subfloat[ \label{fig:rgb-over-fish}]{\includegraphics[width=0.24\textwidth]{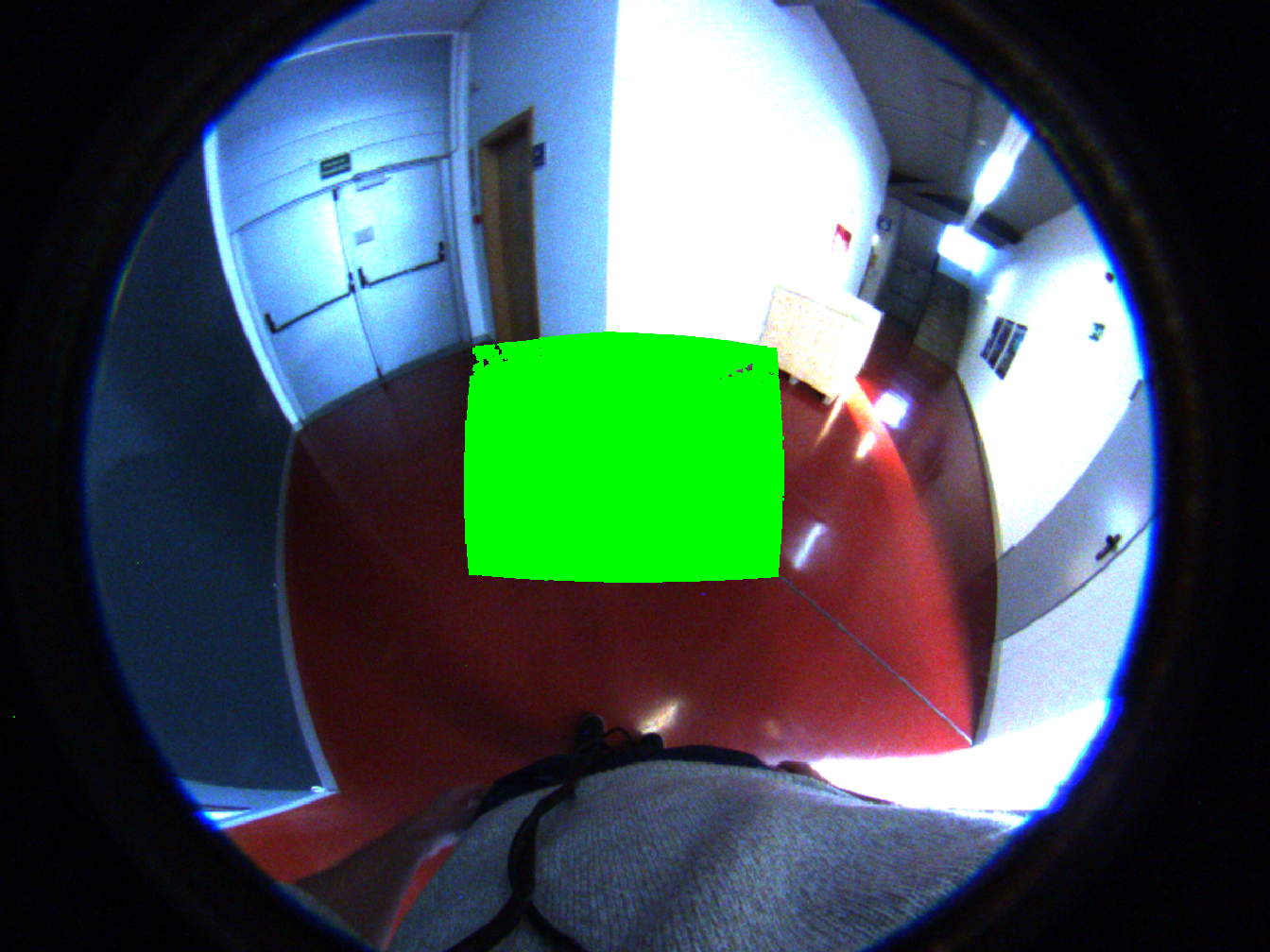}}
    \hfill
    \subfloat[ \label{fig:floor-fish}]{\includegraphics[width=0.24\textwidth]{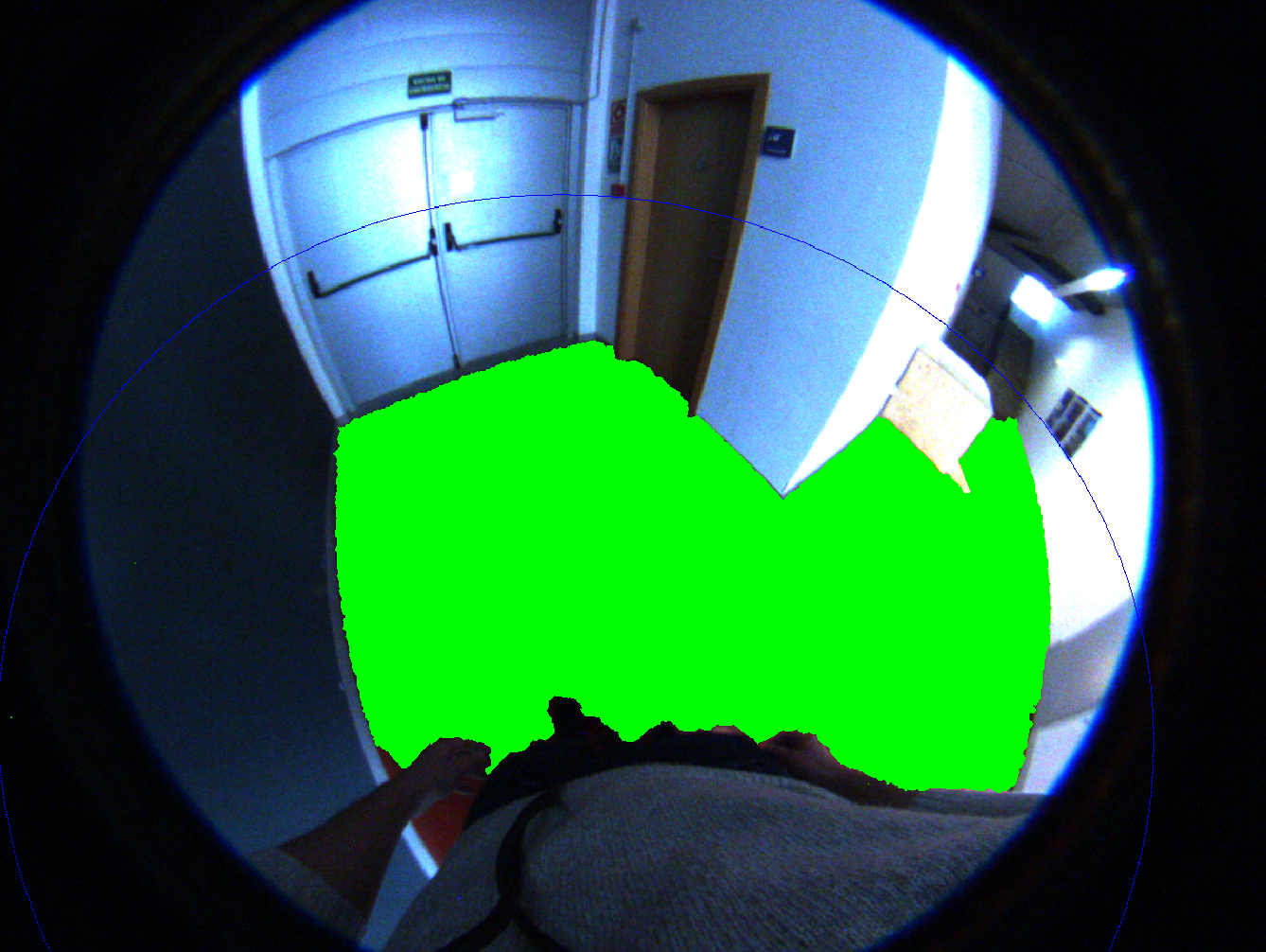}}
    \caption{(a): Proposed combination of cameras; (b): View from a standard RGB-D camera; (c): Projection of the RGB-D field of view into the fish-eye image; (d): Floor detection from the combination of cameras proposed.}
    \label{fig:opening}
\end{figure}

Due to the decrease of prices and size of cameras, they have become the main sensor in new environment perception and navigation systems. Current cameras also have great portability and are small enough to be easily attached to wearable devices. Besides, these cameras provide a great amount of information that can be processed quickly with image processing algorithms. Many recent works use these cameras for helping visually impaired people to avoid obstacles or move in an environment. On \cite{yang2018long,han2015fuzzy,zhang2015slam}, a system using RGB-D cameras and IMU sensors are combined to detect objects at a close range. In \cite{bai2017smart}, authors use an RGB-D camera and an ultrasonic sensor attached to wearable glasses in order to detect transparent and small objects. Augmented reality is used to guide visually impaired people while voice commands are used for totally blind users. In other approach with cameras, \cite{chen2017ccny} take advantage of SLAM information given by a Google Tango device to create safe paths in indoor environments. A control panel is mounted on a white cane that enables the user to communicate with the navigation software via haptics. Besides, different communicating methods with the user are implemented in the state of the art: haptics as vibrotactile commands \cite{lee2016rgb}; image information as augmented reality systems \cite{bai2017smart} or audio with bone-conduction headphones \cite{yang2018long}. Different systems of indoor navigation systems can be found in \cite{plikynas2020indoor}.

The processing of 3D information with RGB-D cameras is a rising approach where many systems for obstacle avoidance have appeared with really good results. Works as \cite{zeineldin2016fast,peasley2013real,wang2014segment,yang2016expanding,pham2016real} use commercial RGB-D cameras to extract the floor surface of the scenes where the user, either blind people or robots, can move safely. However, this kind of system is constrained by the short range and narrow field of view of RGB-D cameras, not being able to obtain a general view of the scene where the navigation is proposed.

\begin{figure}[!t]
	\centering
	\subfloat[]{\includegraphics[width=0.24\textwidth]{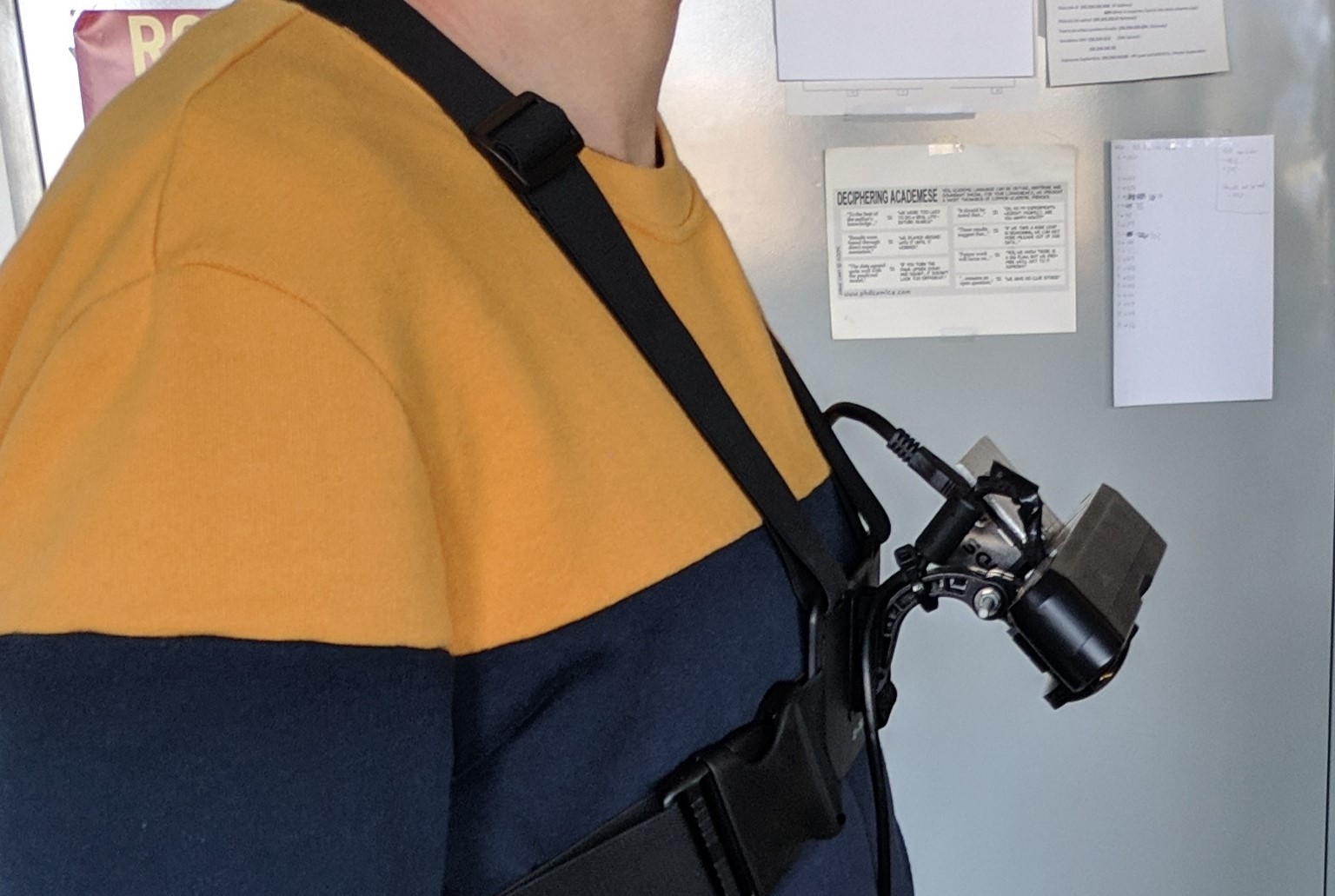} \label{fig:pos_sys}}
	\hfil
	\subfloat[]{\includegraphics[width=0.23\textwidth]{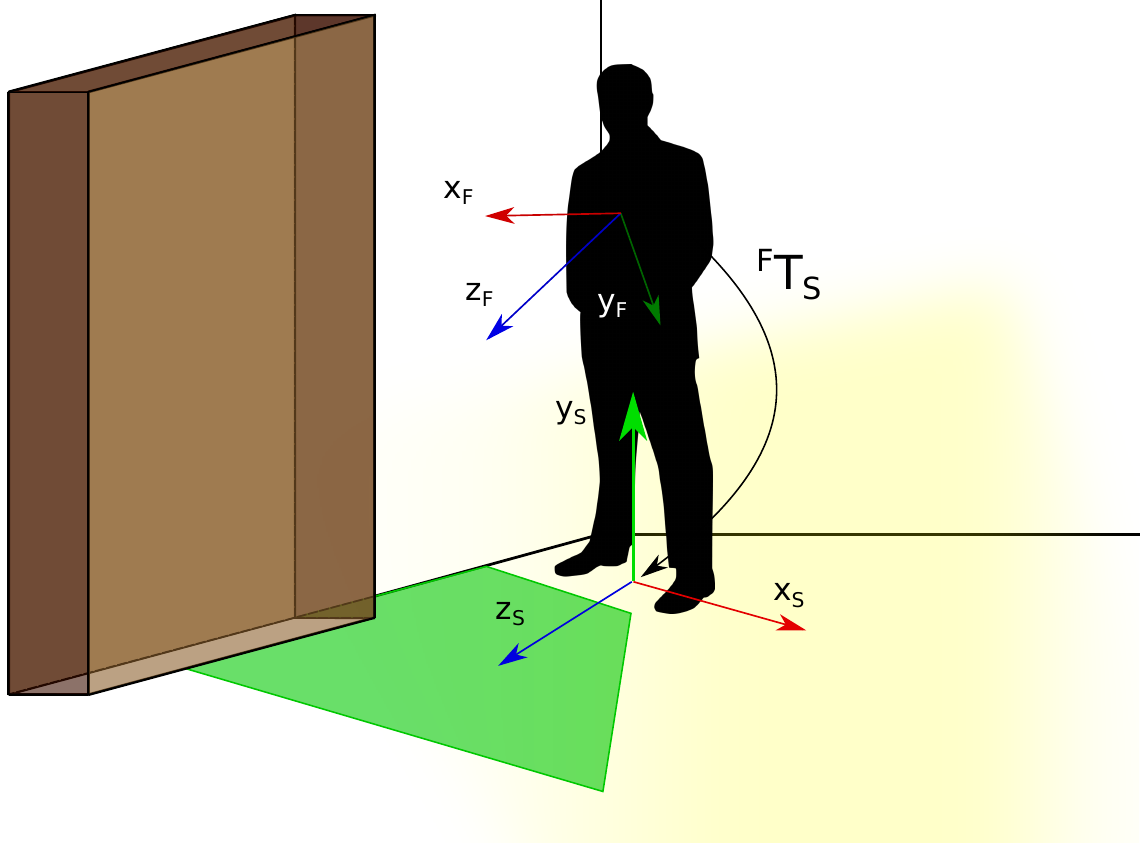} \label{fig:cam2floor}}
	\caption{(a): Proposed device mounted on the user; (b): Reference systems for the user and the camera.}
	
\end{figure}

This work is based in a previous navigation assistance system \cite{aladren2014navigation} which is focused on obstacle avoidance using an RGB-D camera and audio commands. However, inspired by the work \cite{perez2016peripheral}, this new proposal combines the RGB-D camera with a fish-eye camera in order to obtain a better understanding of the scene. With this combination of sensors (see Fig. \ref{fig:rgb-fish}), the extraction of obstacle-free zones in the environment is achieved, allowing to plan safe trajectories instead of just avoiding the obstacles which are close to the user. 
The proposed device takes advantage of the robustness given by the RGB-D cameras, which provide accurate measures of the environment, while a fish-eye camera is used to obtain as much relevant information of the scene as possible, where the wider field of view allows to get a general view of the environment on each frame.
The combination of sensors is mounted on a wearable device (see Fig. \ref{fig:pos_sys}), allowing the user to move freely in any environment.

\section{Background}
\label{sec:background}

Fast development in computer vision has increased the optical devices in the market, reducing prices of these systems. RGB-D cameras are a good example of systems that can simultaneously obtain depth and colour information at an affordable price. However, current RGB-D devices are limited in range and field of view (see Fig. \ref{fig:rgb-fov}), constraining the functionality of the systems they are mounted on. 
For guiding visually impaired in unknown environments, not only avoiding obstacles, a wider perception of the environment is needed as well as the depth information given by these cameras. In order to enhance the functionality of the system, a fish-eye camera is attached to the device to obtain colour images with a larger field of view (see Fig. \ref{fig:rgb-over-fish}). Although these lenses introduce heavy distortions in the image, more information can be obtained as well as a better understanding of the scene than with conventional cameras. Combining these cameras in one device (see Fig. \ref{fig:rgb-fish}), allows to get the better of both: an accurate 3D information from the RGB-D camera, which provides robustness to the navigation system, and a wide field of view from the fish-eye camera, which allows to extend the 3D information from the depth camera up to 20 times in the scene (see Fig. \ref{fig:floor-fish}).

Since the proposed hybrid system has two different cameras \cite{perez2016novel}, a method to calibrate them is needed. The intrinsic parameters of the camera have to be known as well as the extrinsic parameters, which depend on how the system is mounted. The RGB-D camera is modeled as a perspective camera since it does not present big distortions. Since it is a known projection model, it is easier to obtain the calibration parameters and use the intrinsic parameters provided by the manufacturer. However, the projection model of the fish-eye camera is not known. In order to get the calibration parameters, the Scaramuzza's model for revolution symmetry systems \cite{scaramuzza2006toolbox} is used. This method allows to obtain an empiric projection model for fish-eye and other omnidirectional cameras.

\begin{figure*}[!t]
    \centering
    \includegraphics[width = \textwidth]{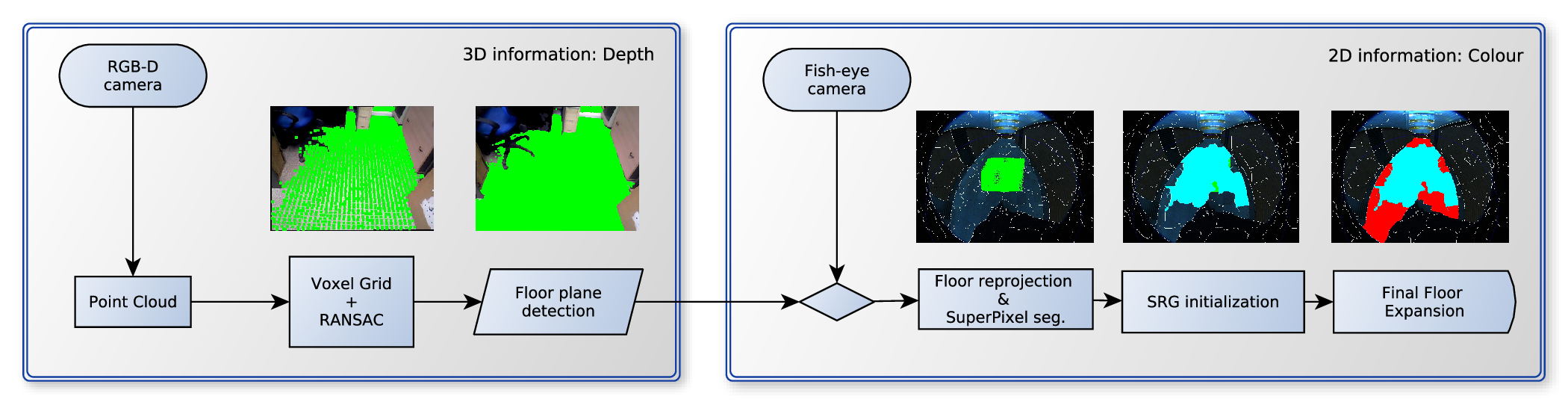}
    \caption{Stages on the floor detection. Starting from the RGB-D camera, where 3D information is obtained from point clouds to extract the floor plane, to the fish-eye image, where working with colour information in order to extend the floor plane in a wider range into the environment.}
    \label{fig:diagram}
\end{figure*}

The extrinsic calibration parameters between the depth and fish-eye cameras are obtained with the method defined in \cite{perez2017extrinsic}. With this method, lines from the scene are extracted and matched in both RGB-D and fish-eye camera. From the RGB-D camera, the line extraction is obtained as plane intersections in the 3D space, with only depth information, or using a segment extraction in the RGB image and obtaining the 3D line from the depth information. From the RGB-D camera a 3D point and direction are computed for each extracted line, defined as $\mathbf{p_i}^D$ and $\mathbf{v_i}^D$ respectively for a line $i$. Since fish-eye cameras present heavy distortions, the method presented in \cite{bermudez2015automatic} for the line extraction is used. From these lines, the normal vector of the plane that contains the line and the optical center is obtained, defined as $\mathbf{n_i}^F$ for a line $i$. Line correspondences are made fusing the information obtained from the RGB-D camera and the fish-eye as presented in \cite{perez2017extrinsic}. Every correspondence is defined as $L_i = \{ \mathbf{p_i}^D,\mathbf{v_i}^D,\mathbf{n_i}^F \}$, where the parameters must fulfill the equation of the plane in which the line is contained. From the line correspondences, the extrinsic parameters are obtained solving a non-linear least squares problem with the Gauss-Newton method. From equation \eqref{eq:R} is computed the relative rotation between the cameras and from equation \eqref{eq:t} the translation.

\begin{equation}
    \underset{\mu}{\arg\min} \sum_{i=1}^{N_L}( \mathbf{n_i}^T \cdot e^\mu \mathsf{R} \mathbf{v_i})^2
    \label{eq:R}
\end{equation}

\begin{equation}
    \underset{\mathbf{t}}{\arg\min} \sum_{i=1}^{N_L}\left(\mathbf{n_i}^T \cdot \dfrac{\mathsf{R} \mathbf{p_i} + \mathbf{t}}{\left \| \mathsf{R} \mathbf{p_i} + \mathbf{t} \right \|}\right)^2
    \label{eq:t}
\end{equation} \noindent
where $\mathsf{R}$ and $\mathbf{t}$ are the extrinsic parameters of calibration, $e^\mu$ is the exponential map of the increment of rotation $\mu$ on $\mathsf{R}$ and $N_L$ is the number of line correspondences.

\section{Perception system}
\label{sec:method}

The navigation system should be able to guide the user through the obstacle-free zones of the scene. This method proposes to extract the floor plane from the 3D information given by the RGB-D camera, concealed by a small field of view. Then, the obstacle-free floor zones are expanded with the colour information given by the wide field of view of the fish-eye camera. This combination of sensors, as detailed in \cite{perez2016peripheral,perez2016novel,perez2019scaled}, allows to obtain accurate 3D information from a small part of the scene and extend this information to a larger part of the environment. Furthermore, using the same combination of cameras, a new detection and localization system for doors is proposed, so the user would be able to continue the navigation through separated environments.

In the proposed method, the scene's floor recovery is made in two stages (see Fig. \ref{fig:diagram}). In the first stage, depth information is used to extract the floor with the RGB-D camera, which gives 3D information of the plane in which the floor is contained. On the second stage, colour information is used to extend the floor plane taking advantage of the wider field of view of the fish-eye camera. In order to combine both stages, the information from the RGB-D camera is transformed into the fish-eye reference through the extrinsic parameters computed in section \ref{sec:background}.

\subsection{Floor detection}
\label{subsec:floor-det}

\begin{figure*}[!t]
    \centering
    \subfloat[]{\includegraphics[width = 0.22\textwidth]{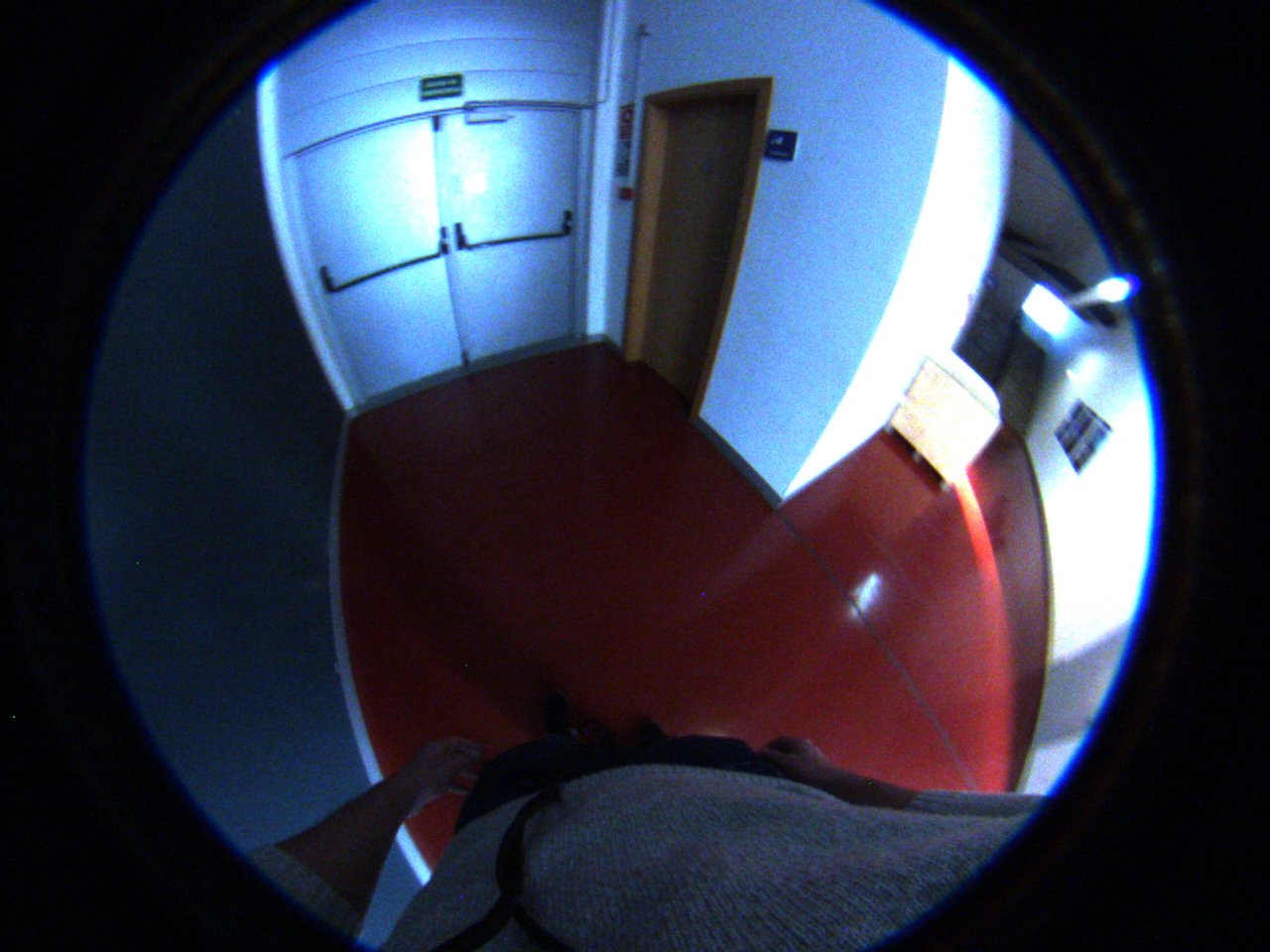} \label{fig:fish-aq}}
    \hfil
    \subfloat[]{\includegraphics[width = 0.22\textwidth]{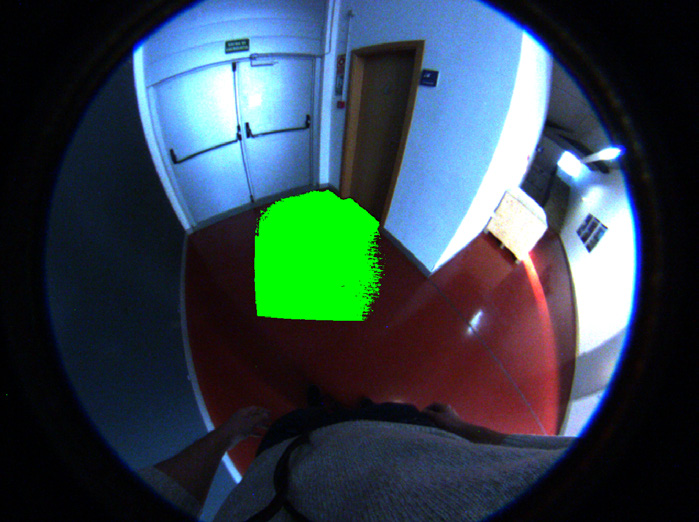} \label{fig:floor-rgbd}}
    \hfil
    \subfloat[]{\includegraphics[width = 0.22\textwidth]{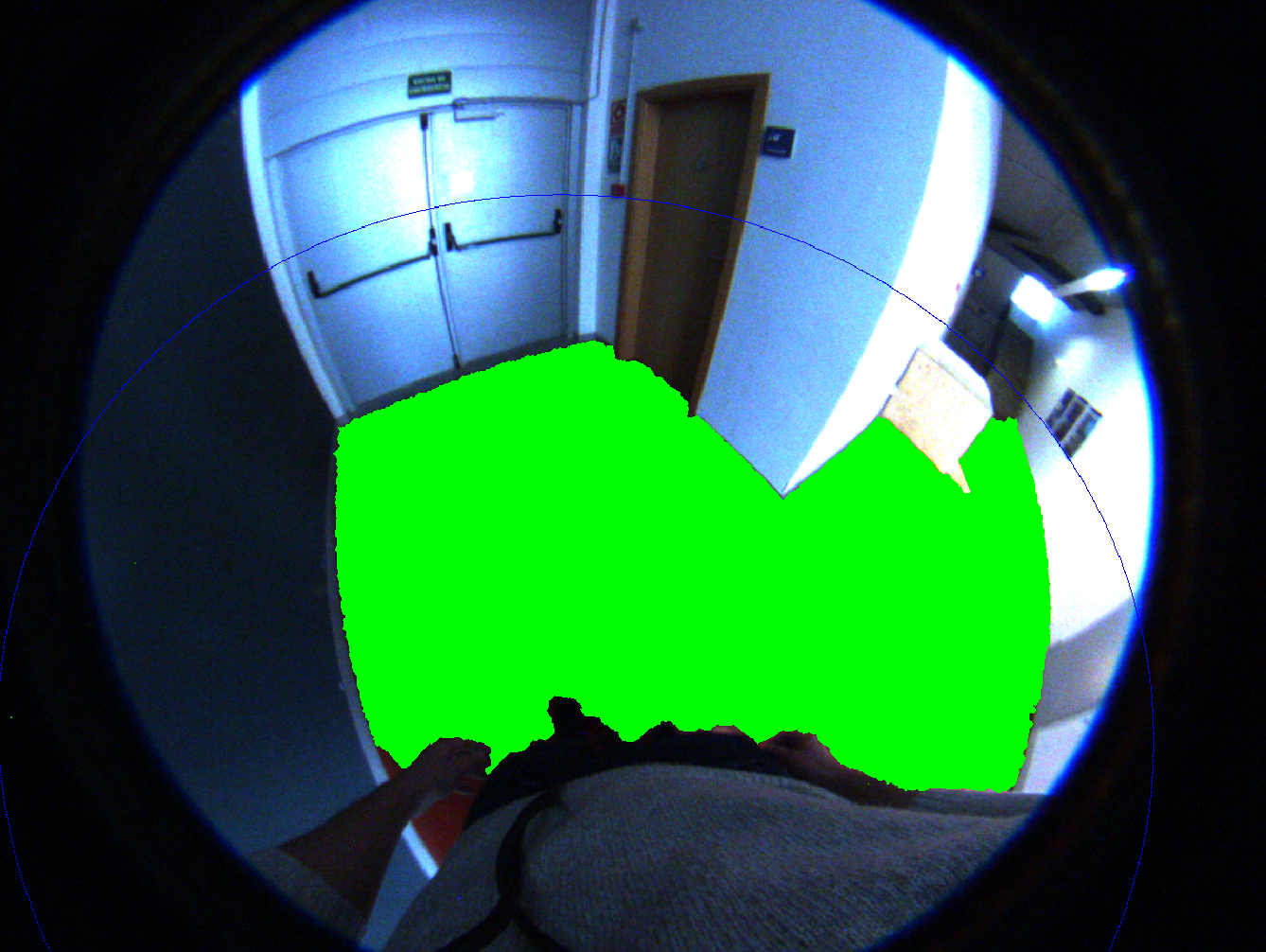} \label{fig:floor-exp}}
    \hfil
    \subfloat[]{\includegraphics[width = 0.3\textwidth]{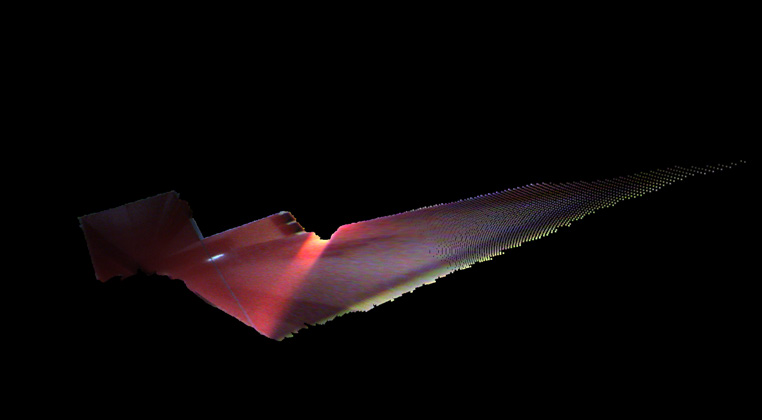} \label{fig:3D-rec}}
    \caption{(a): Fish-eye image; (b): RGB-D floor detection; (c): Floor expansion; (d): 3D reconstruction of the obstacle-free zones.}
    \label{fig:in-c1}
\end{figure*}

On the first stage, to obtain the 3D plane that contains the floor of the scene, point clouds given by the RGB-D camera are used. In order to speed up the implementation, the point cloud is downsampled into a voxel grid, reducing the noise and smoothing the surfaces as side effect. Taking the information from the voxel grid as input, a RANSAC algorithm is used to obtain the equation of the main plane in the image. Due to the orientation of the camera, the dominant plane in the image will be the floor plane. Once the equation of the floor plane, $n_x X + n_y Y + n_z Z + n_0 = 0$, and its normal vector, $\mathbf{n} = (n_x,n_y,n_z)^T$ are obtained, the points from the point cloud that belong to the floor plane can be defined. Those points within a close distance to the plane are considered as points belonging to the floor, while the rest of the points are considered as not floor or obstacles.

After obtaining the floor plane, the information given by the RGB-D camera is transformed into the fish-eye camera. For this purpose, the extrinsic parameters computed in section \ref{sec:background} are used. With this transformation, it is posible to obtain the floor plane information in the fish-eye reference, where the method will take advantage of the wide field of view to extend the floor plane. The transformation of the points from the RGB-D camera, $\mathbf{p}^D$, into the fish-eye camera, $\mathbf{p}^F$, is computed as:
\begin{equation}
    \begin{pmatrix} \mathbf{p}^F \\ 1  \end{pmatrix} =
    \begin{pmatrix} \mathsf{R} & \mathbf{t} \\ 0 & 1  \end{pmatrix} \cdot
    \begin{pmatrix} \mathbf{p}^D \\ 1  \end{pmatrix}
    \label{eq:reproject}
\end{equation} \noindent
where $\mathsf{R}$ and $\mathbf{t}$ are the rotation and translation obtained from the extrinsic parameters of the cameras.

\subsection{Floor expansion}
\label{subsec:floor-exp}

On the second stage, the colour information from the fish-eye camera is used, in addition to the information from the RGB-D camera. The extension of the floor is made in the colour domain. However, a pixel by pixel comparison would result in a really slow implementation, which is not interesting for a real-time guiding system. So, in order to extend the floor in the colour image, a SuperPixel segmentation \cite{van2012seeds} is used, which group pixels with similar colour into bigger regions. After the Superpixel segmentation, a Seeded Region Growing (SRG) algorithm is used to group SuperPixels. First, the initial seed is set taking the SuperPixels that belong to the reprojection of the extracted floor into the fish-eye image. The colour histograms are computed in the Hue-Saturation colour channels from the initial seed and saved as references. Taking only those two channels allows better segmentation results in scenes with different illumination and surfaces with reflections. By computing the colour histograms from nearby SuperPixels and making the comparison with the reference histograms, the initial seed grows to those zones that have similar colour. The algorithm stops when the surrounding SuperPixels have colour histograms different from the reference or when they are over the horizon line, which works as an upper limit.
The horizon line is computed as those projecting rays parallel to the floor plane. That is the same as obtaining the rays that fulfil: $\mathbf{v}^F\cdot \mathbf{n} = 0$, where $\mathbf{v}^F$ is a projecting ray of the fish-eye camera and $\mathbf{n}$ is the normal vector of the floor plane.

\subsection{Door detection}
\label{subsec:door}

By contrast with previous approaches using conventional and stereo cameras for door detection \cite{murillo2008visual}, the proposed combination of cameras takes advantage of the previous floor extraction and expansion. The fish-eye image is used to generate hypotheses of possible doors in the scene. Taking advantage of the segmented floor, measurements in real dimensions can be made in order to verify these hypotheses.

\begin{figure*}[!t]
	\centering
	\subfloat[]{\includegraphics[width = 0.24\textwidth]{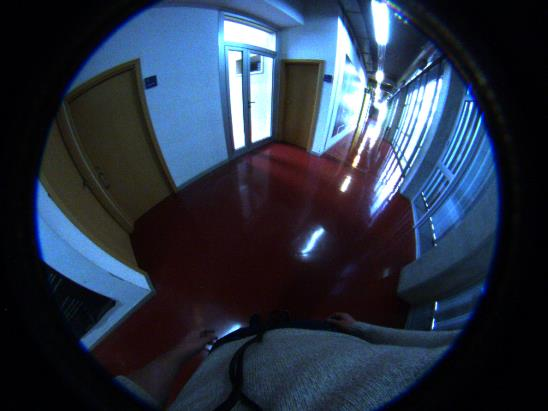} \label{fig:door-fish}}
	\hfil
	\subfloat[]{\includegraphics[width = 0.24\textwidth]{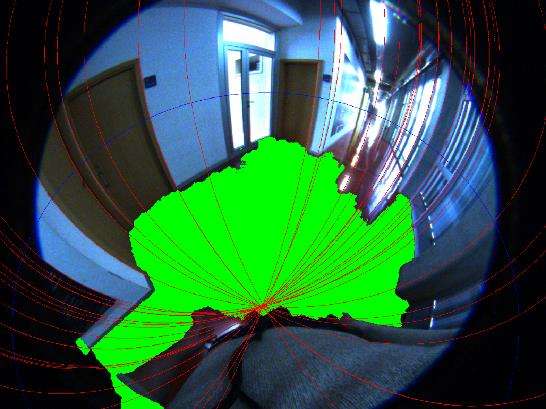}\label{fig:door-lines}}
	\hfil
	\subfloat[]{\includegraphics[width = 0.24\textwidth]{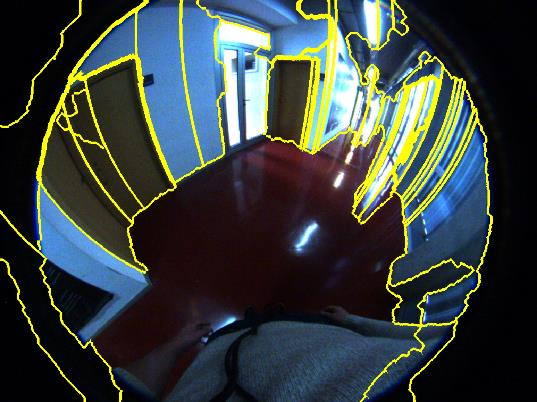}\label{fig:door-hypo}}
	\hfil
	\subfloat[]{\includegraphics[width = 0.24\textwidth]{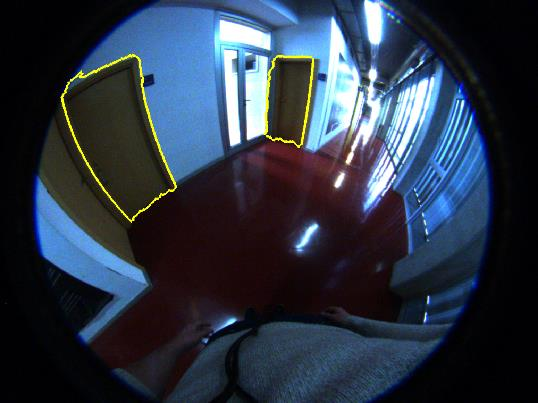}\label{fig:door-result}}
	\caption{(a): Fish-eye image; (b): Extended floor and Vertical line extraction; (c): Door hypotheses regions; (d): Door detection.}
	\label{fig:doordetection}
\end{figure*}

For generating hypotheses, the line extraction algorithm for fish-eye cameras described in \cite{bermudez2015automatic} is used. The line extraction is performed and the vertical lines of the scene are defined as those which are perpendicular to the horizon line (see Fig. \ref{fig:door-lines}). With these lines, two kind of door hypotheses are considered. {\it Two-lines} hypothesis: Two consecutive lines separated a distance within a range can form a door; {\it One-line} hypothesis: One line can be part of a door which will be in only one of its sides.

From the segmented floor,  measurements in meters can be obtained. This allows to measure the distance between two lines as the distance between the intersection point of the vertical lines and the extracted floor. Those lines that are 0.6 to 1.4 meters apart are paired and generate a {\it Two-line} hypothesis. This range is defined by the dimensions of real doors. From each remaining vertical line, two {\it One-line} hypotheses are generated, one at each side of the line. The door hypothesis is completed with an uniform colour region, which is mostly different from the colour of walls and floor. Thus, taking advantage of the SuperPixel segmentation and SRG algorithm previously developed for the floor expansion, new seeds for each door hypothesis are created. The colour seed is obtained from the SuperPixel in the center of the door hypothesis, obtained with the horizon line. From this algorithm, the region near the lines is expanded, getting regions that define the door hypotheses, as can be seen in Fig. \ref{fig:door-hypo}.

In order to verify these hypotheses, first it is considered the real height of the doors by computing a Cross-Ratio, CR. Due to the invariant nature of the CR, the real height of doors can be measured from its projection in the fish-eye image. To define the CR, four points are taken (A,B,C,D) along each vertical line, which are its intersection with: the segmented floor (A), the horizon line (B) and the upper part of the door hypothesis region (C), being point D the vertical vanishing point. With these four points, the CR can be obtained knowing the typical door height (2.00 m) and the height of the camera from the floor $\left (  n_0/\sqrt{{n_x}^2+{n_y}^2+{n_z}^2} \right )$, since 

\begin{equation}
    CR = \dfrac{AC \cdot BD}{AD \cdot BC} = \dfrac{AC}{BC}
\end{equation} \noindent
when D is at infinity. By computing the CR for each hypothesis, those that are within a threshold from the standard value are kept.

A second constraint is the distance from the coloured region to the floor plane. Ideally this distance should be zero for doors and bigger for other door-like objects as windows or frames. So, those door hypothesis which distance to the floor plane is greater than a threshold distance are rejected. The {\it Two-line} hypotheses that fulfil these two constraints are verified as doors in the scene. For the {\it One-line} hypothesis, an oriented bounding box (OBB) of the coloured region is obtained. Since the width of the doors cannot be measured from this hypothesis, the aspect ratio of the OBB is computed in order to verify or reject the hypotheses.

On both hypotheses, really restrictive conditions are used to avoid false positives in the door detection. Since one of the priorities is the safety of the guided people, the detection of false doors cannot be allowed. The result of the algorithm can be seen in Fig. \ref{fig:door-result}.

\section{Experiments}
\label{sec:experiments}

A set of experiments are conducted to test the performance of the proposed algorithms for floor extraction and door detection. The chosen environments are zones with different illumination, indoors as well as outdoors.

The proposed system is composed by an RGB-D {\it Asus Xtion Pro Live} camera, with a field of view of 43$\times$57 degrees and $640 \times 480$ pixels resolution, and a camera {\it uEye UI-3580CP}, with resolution $2560 \times 1920$ pixels, carrying a 182 degrees field of view fish-eye lens (Lensagon CF5M1414). For computational purposes, the resolution of the {\it uEye} camera is reduced to $1280 \times 960$ pixels. These two cameras are attached together, see Fig. \ref{fig:rgb-fish}, to maintain the extrinsic calibration parameters fixed. The device is mounted over the chest of the user and oriented to the floor, at about -45 degrees of inclination, as seen in Fig. \ref{fig:pos_sys}. This position gives stability to the image and allows to disregard the small variations on the orientation. Besides, using this configuration, the floor is always kept in the field of view of the RGB-D camera making the floor detection more robust.

\begin{figure*}[th]
\centering

    \subfloat{\includegraphics[width=0.24\textwidth]{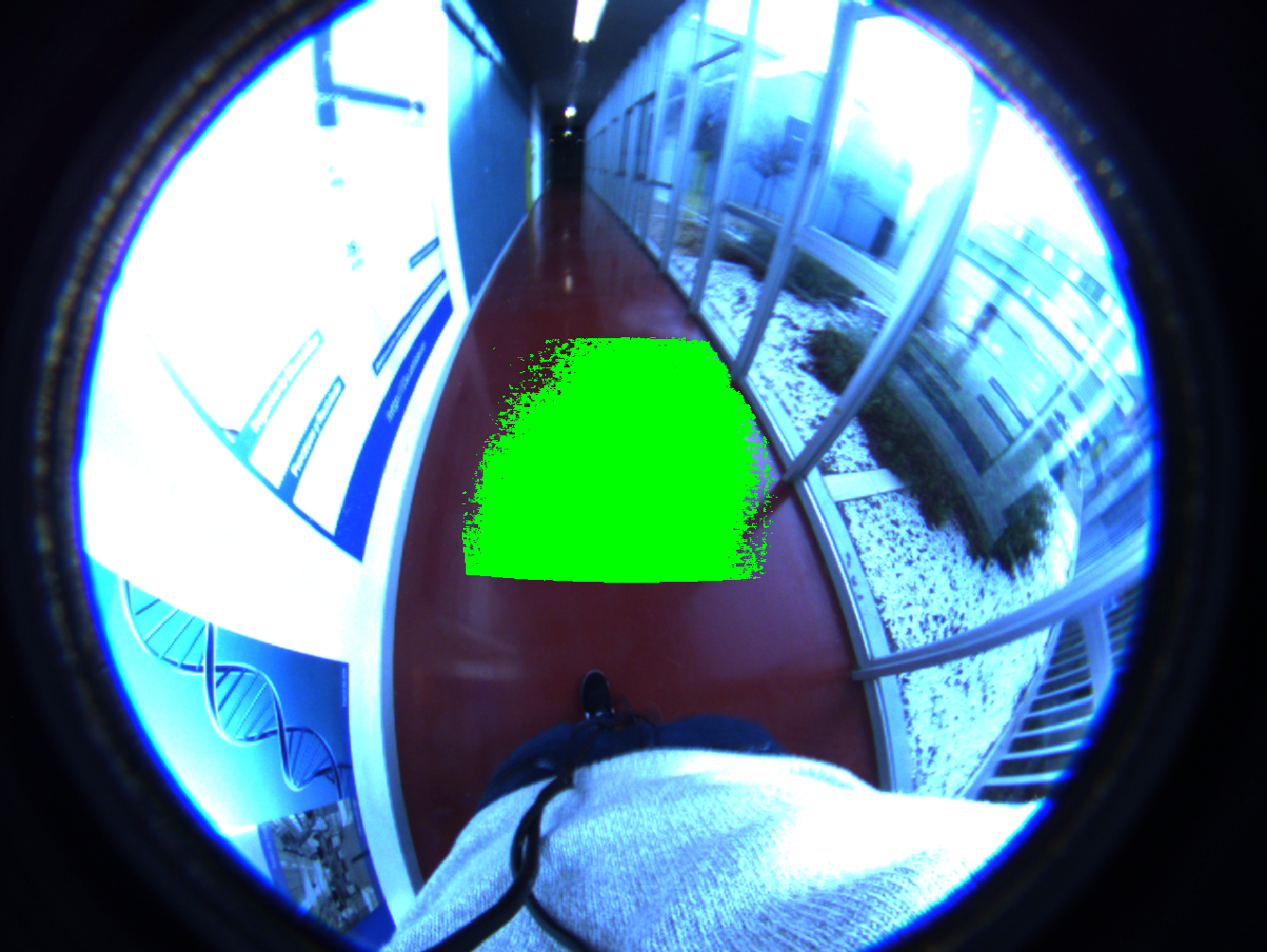}} \hfil
    \subfloat{\includegraphics[width=0.24\textwidth]{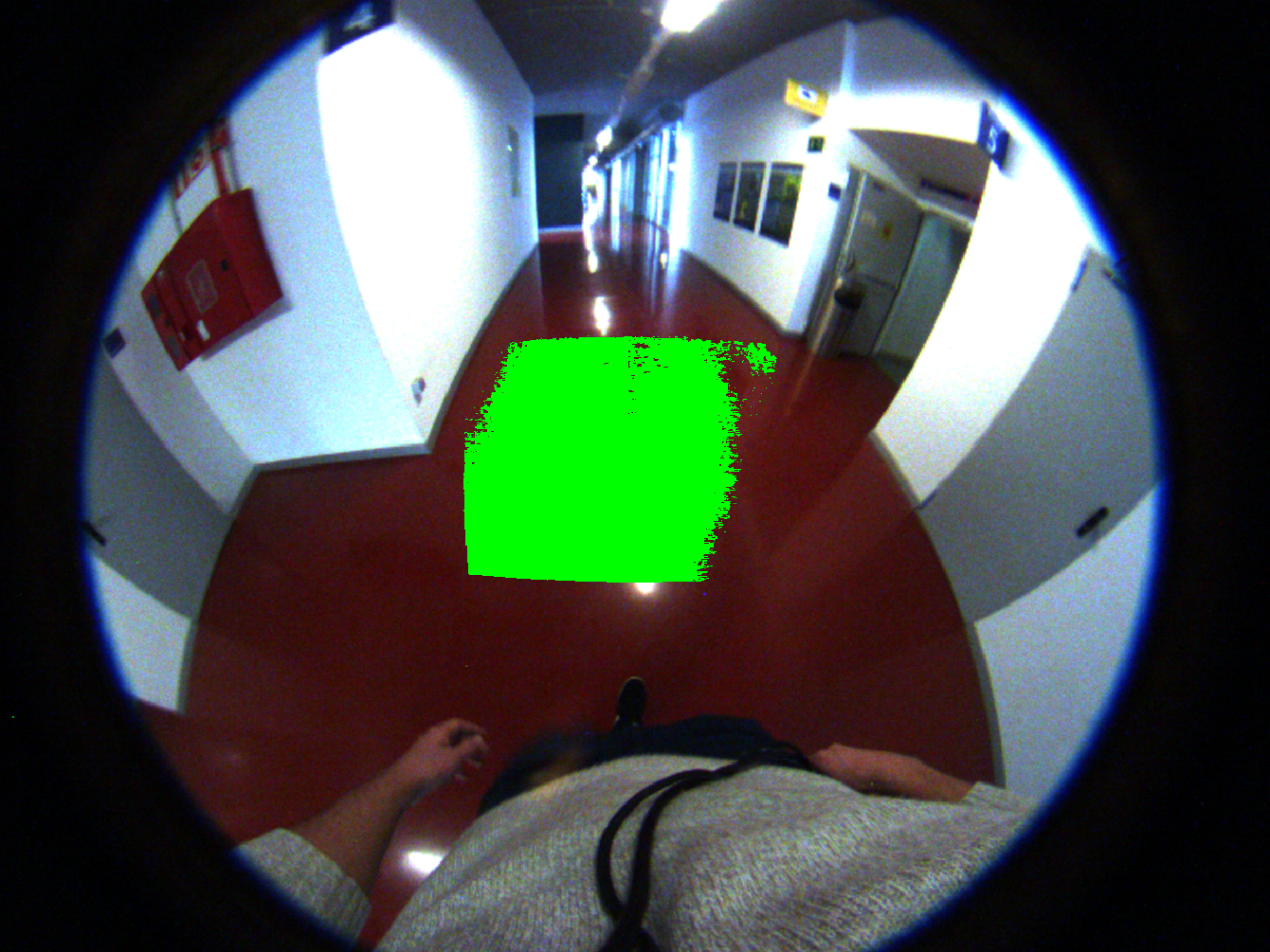}} \hfil  
    \subfloat{\includegraphics[width=0.24\textwidth]{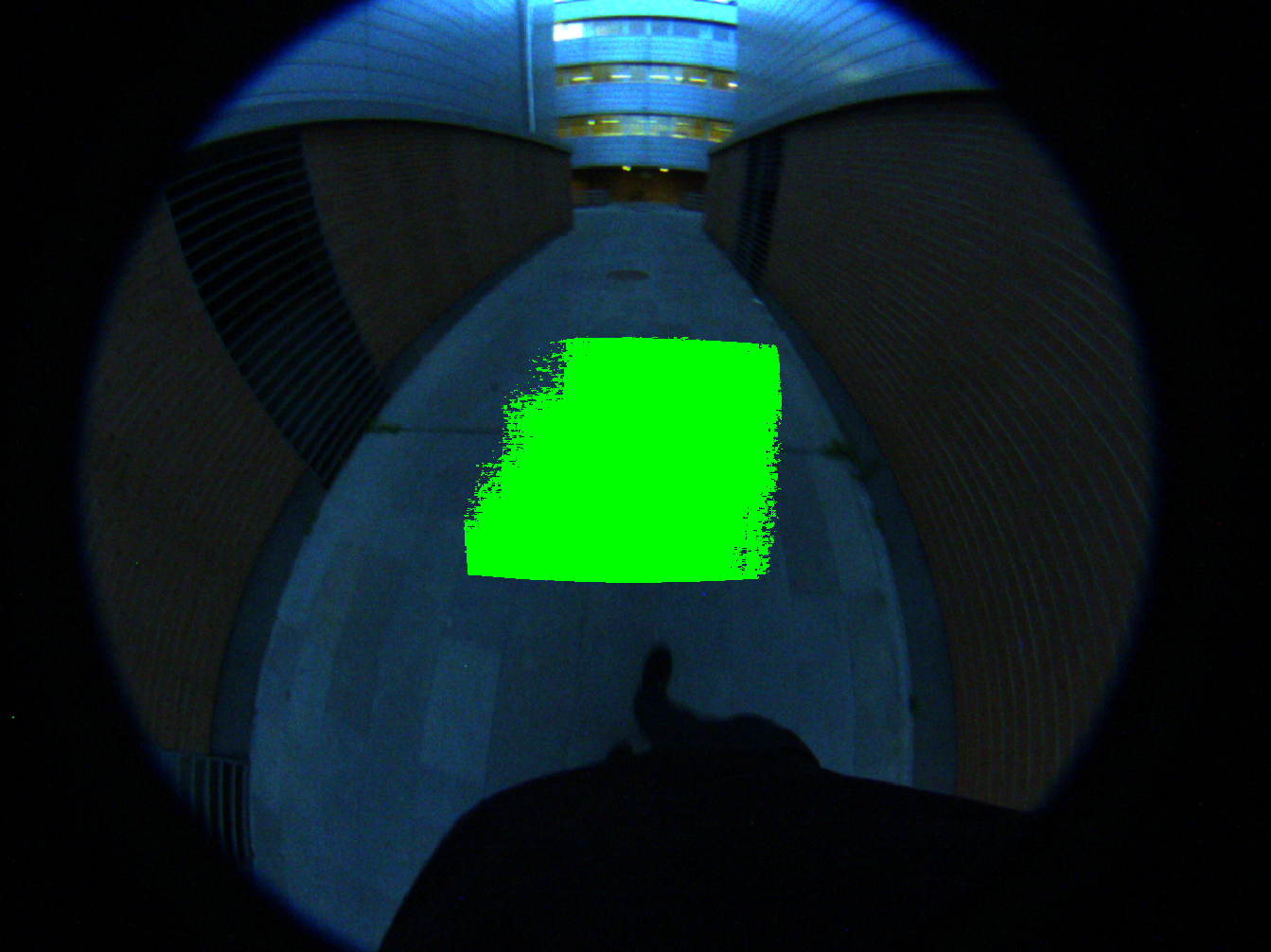}} \hfil  
    \subfloat{\includegraphics[width=0.24\textwidth]{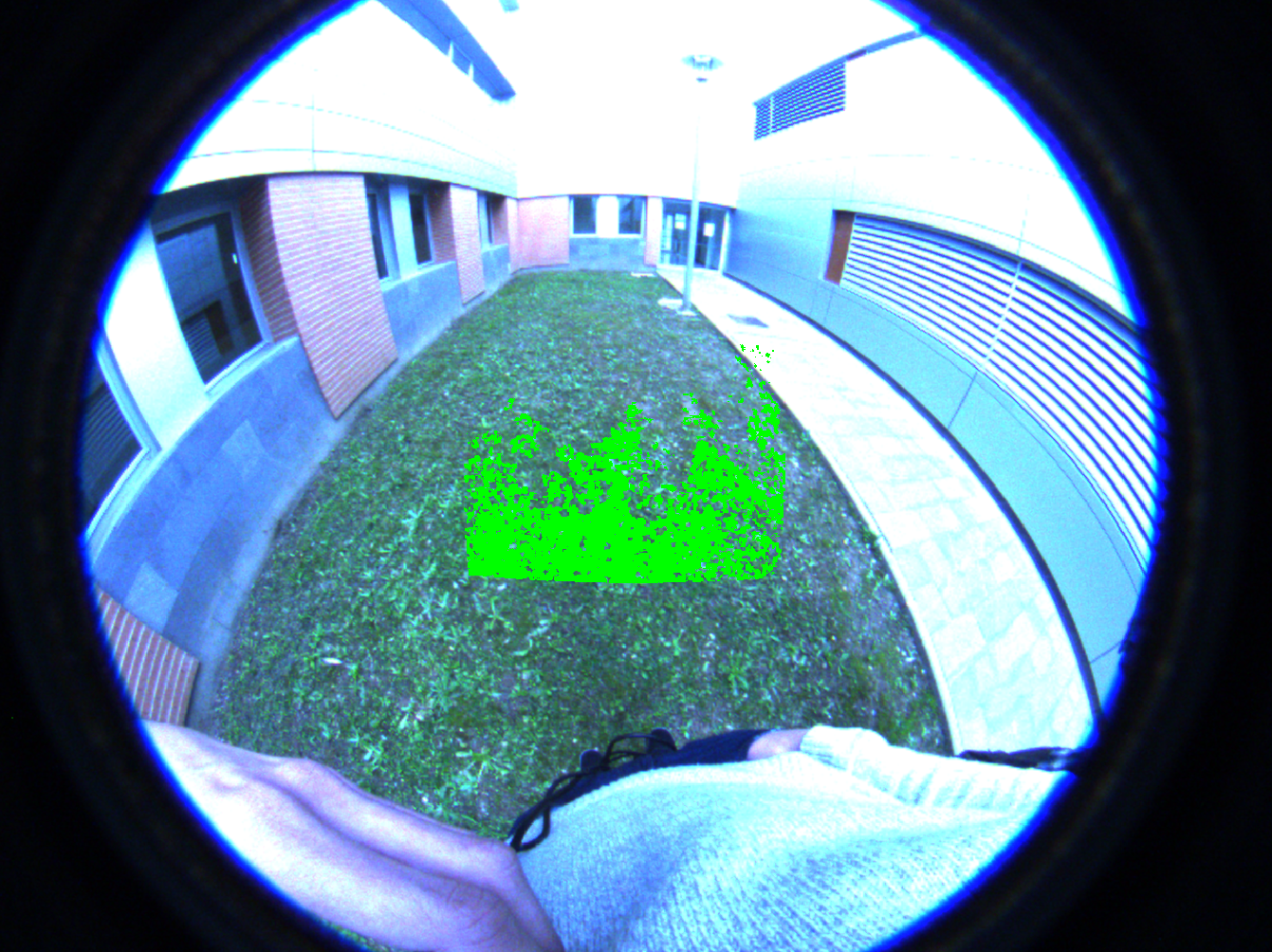}} \\
    \vspace{-2mm}
    \subfloat{\includegraphics[width=0.24\textwidth]{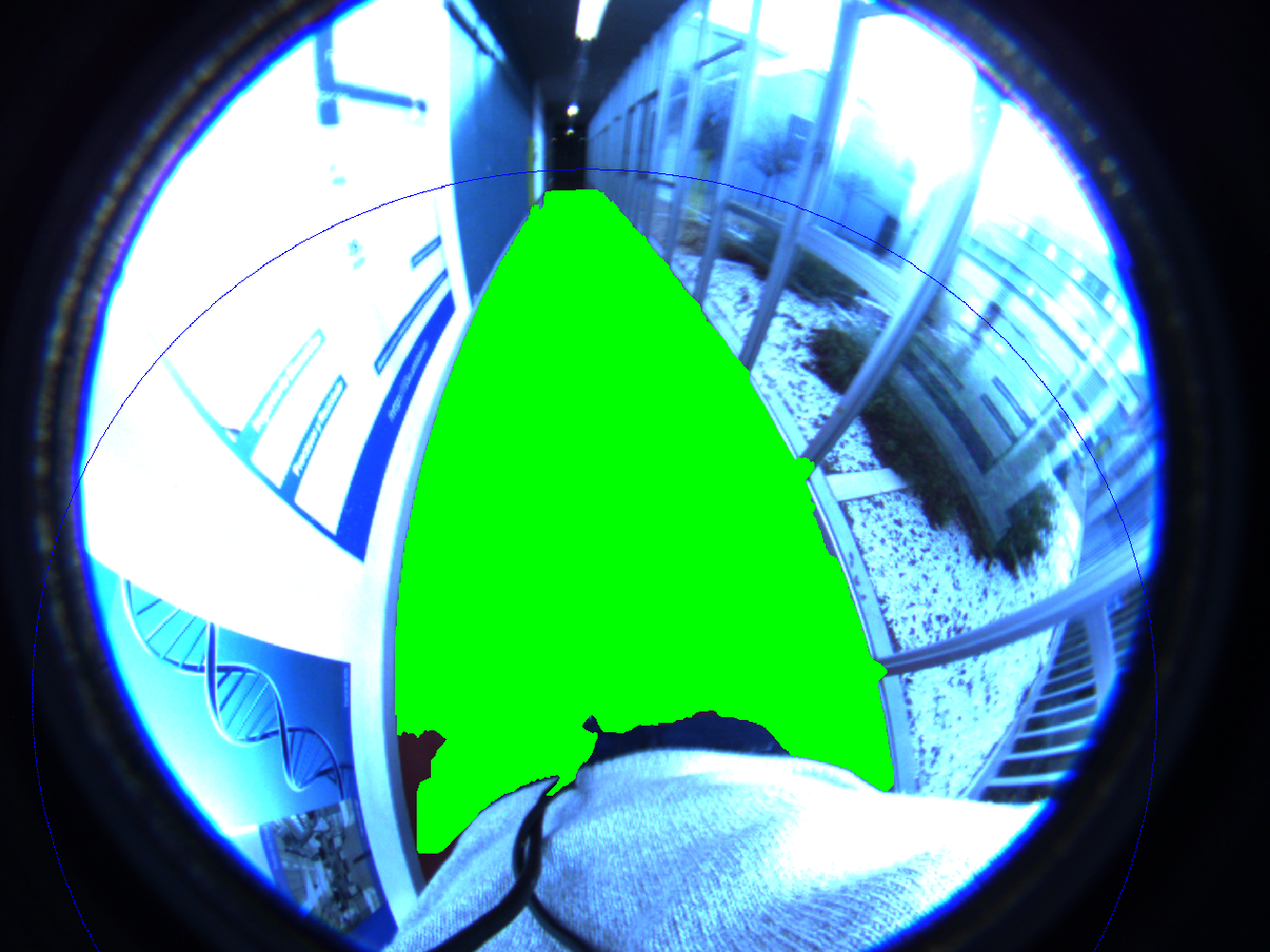}} \hfil
    \subfloat{\includegraphics[width=0.24\textwidth]{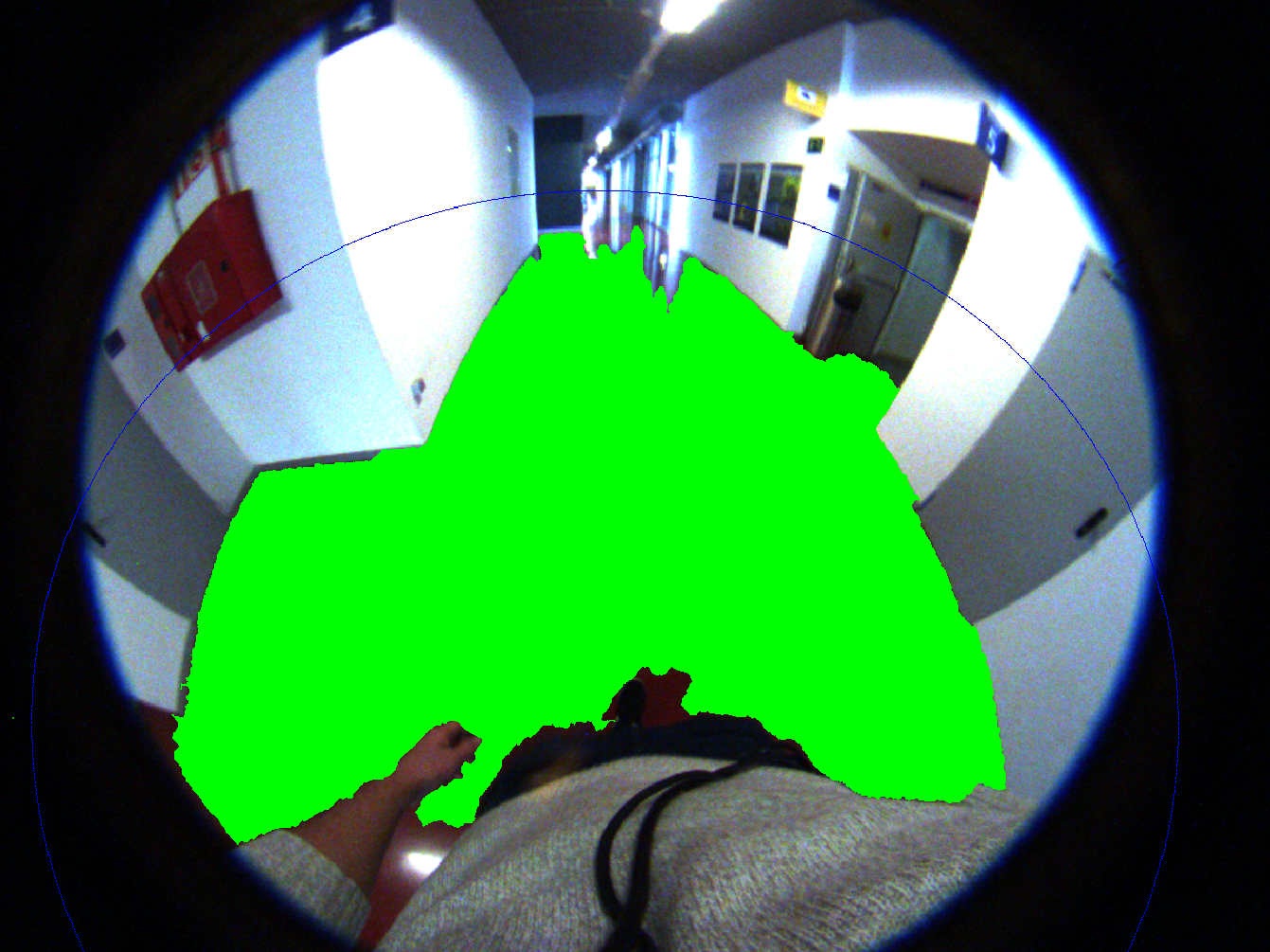}} \hfil
    \subfloat{\includegraphics[width=0.24\textwidth]{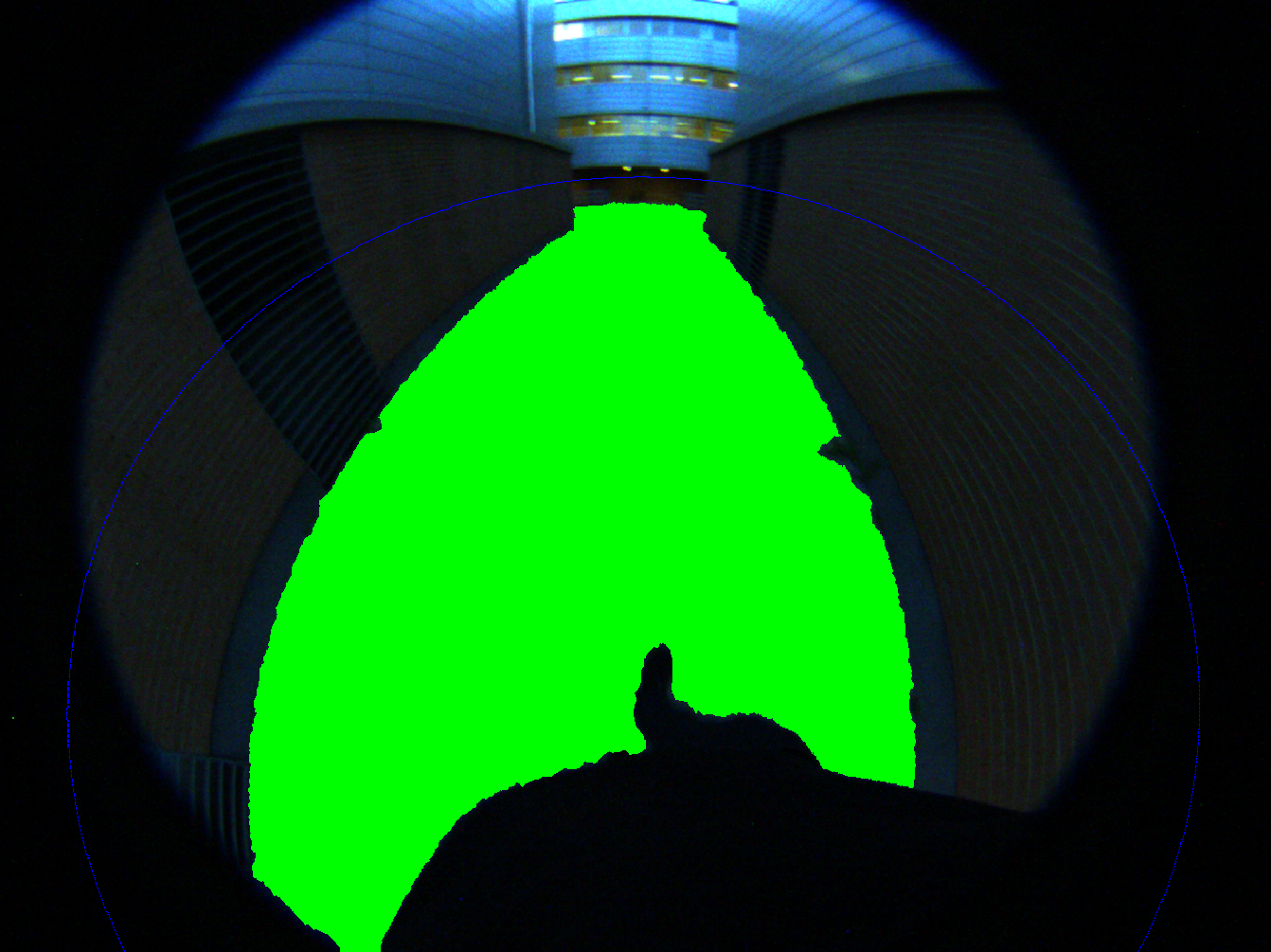}} \hfil
    \subfloat{\includegraphics[width=0.24\textwidth]{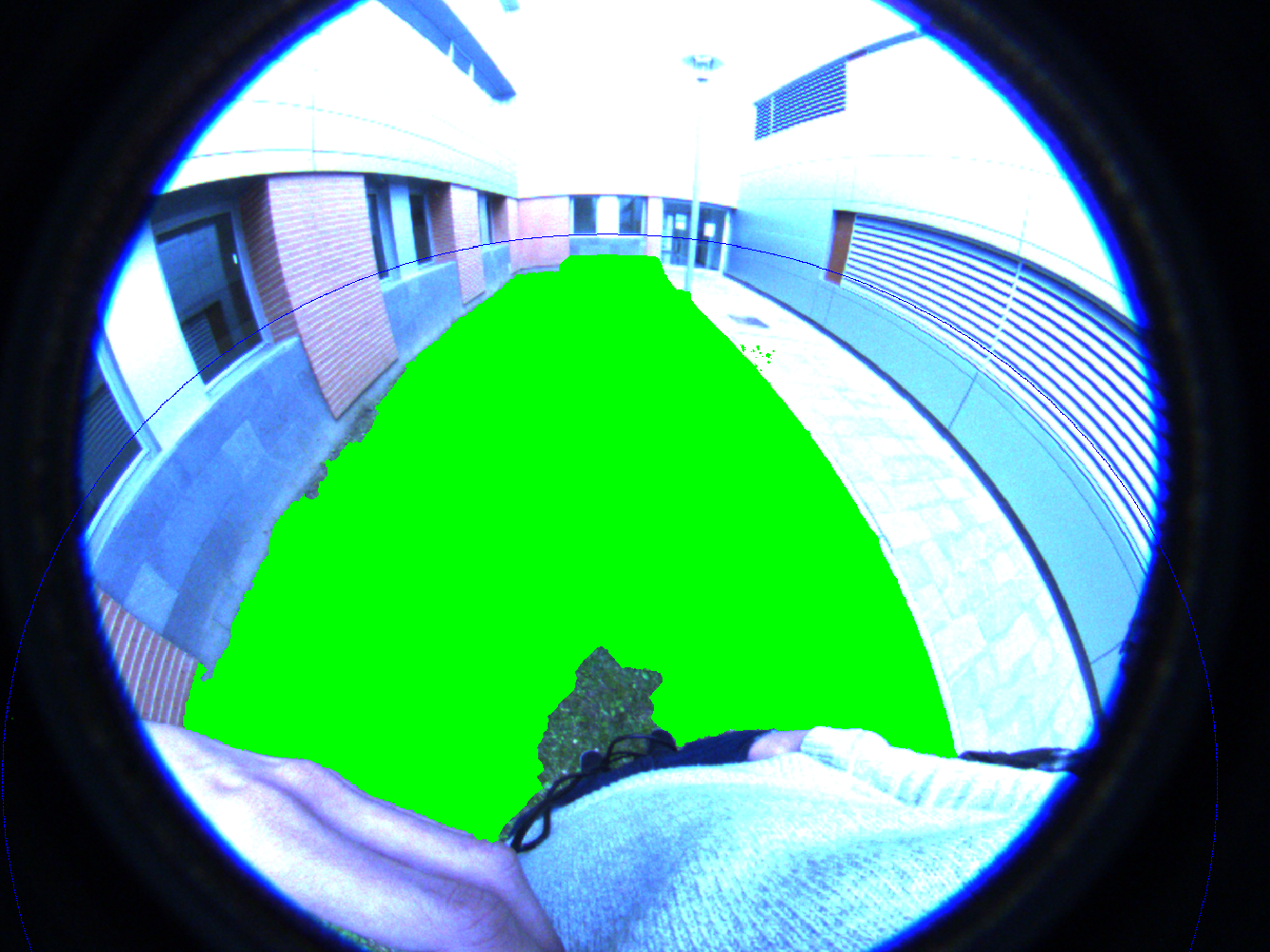}}\\
    \vspace{-2mm}
    \subfloat{\includegraphics[width=0.24\textwidth]{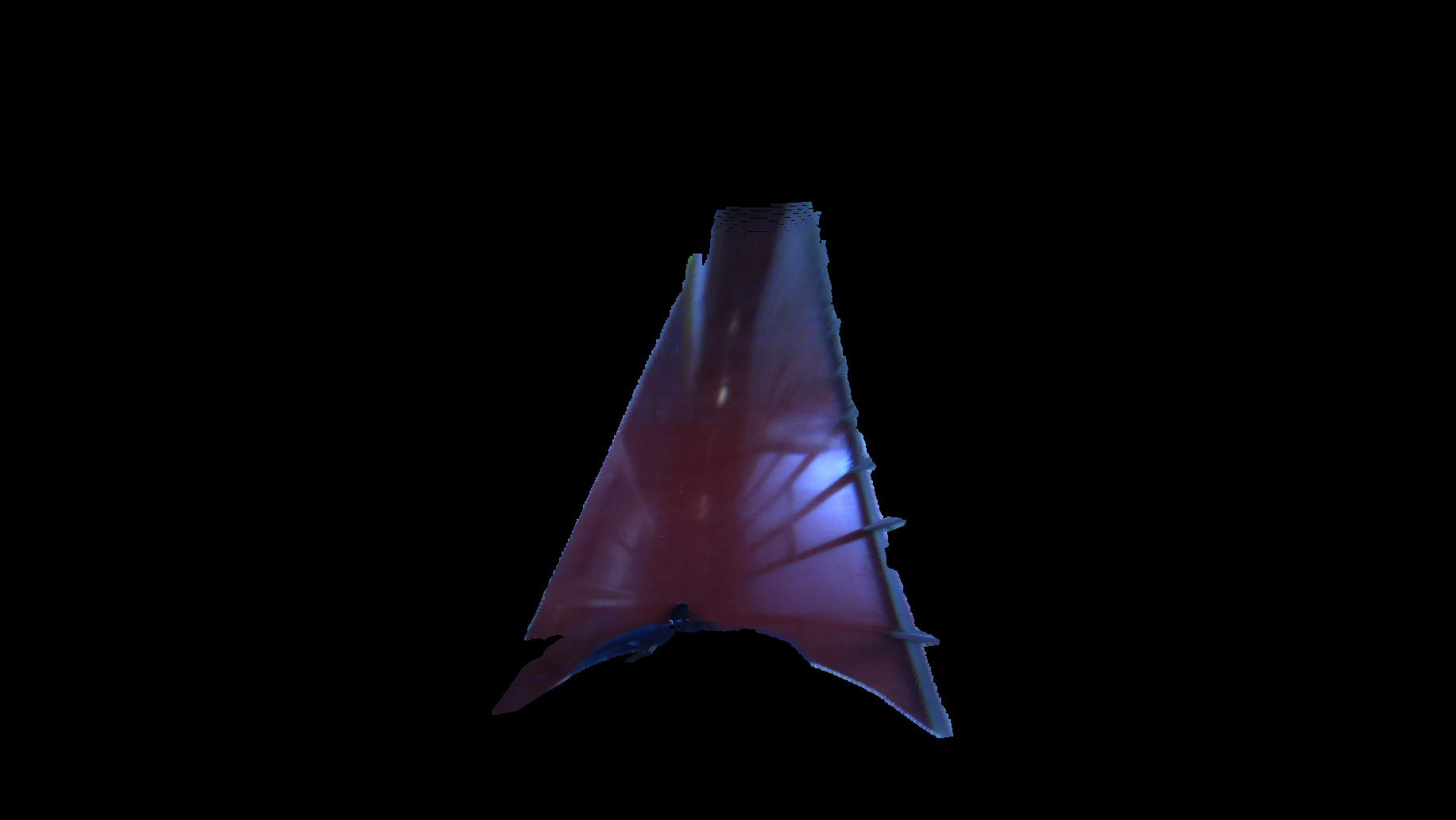}} \hfil    
    \subfloat{\includegraphics[width=0.24\textwidth]{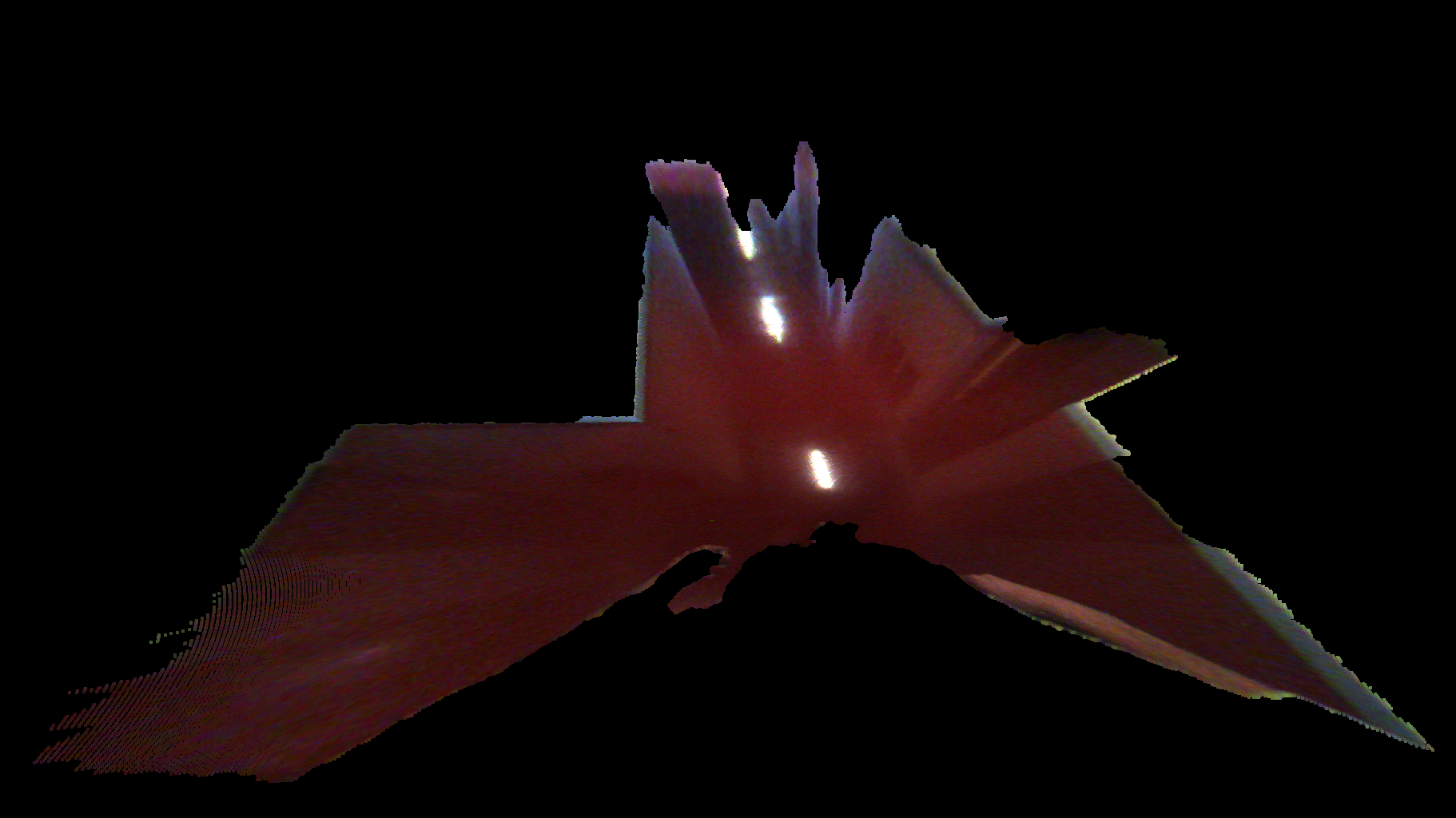}} \hfil        
    \subfloat{\includegraphics[width=0.24\textwidth]{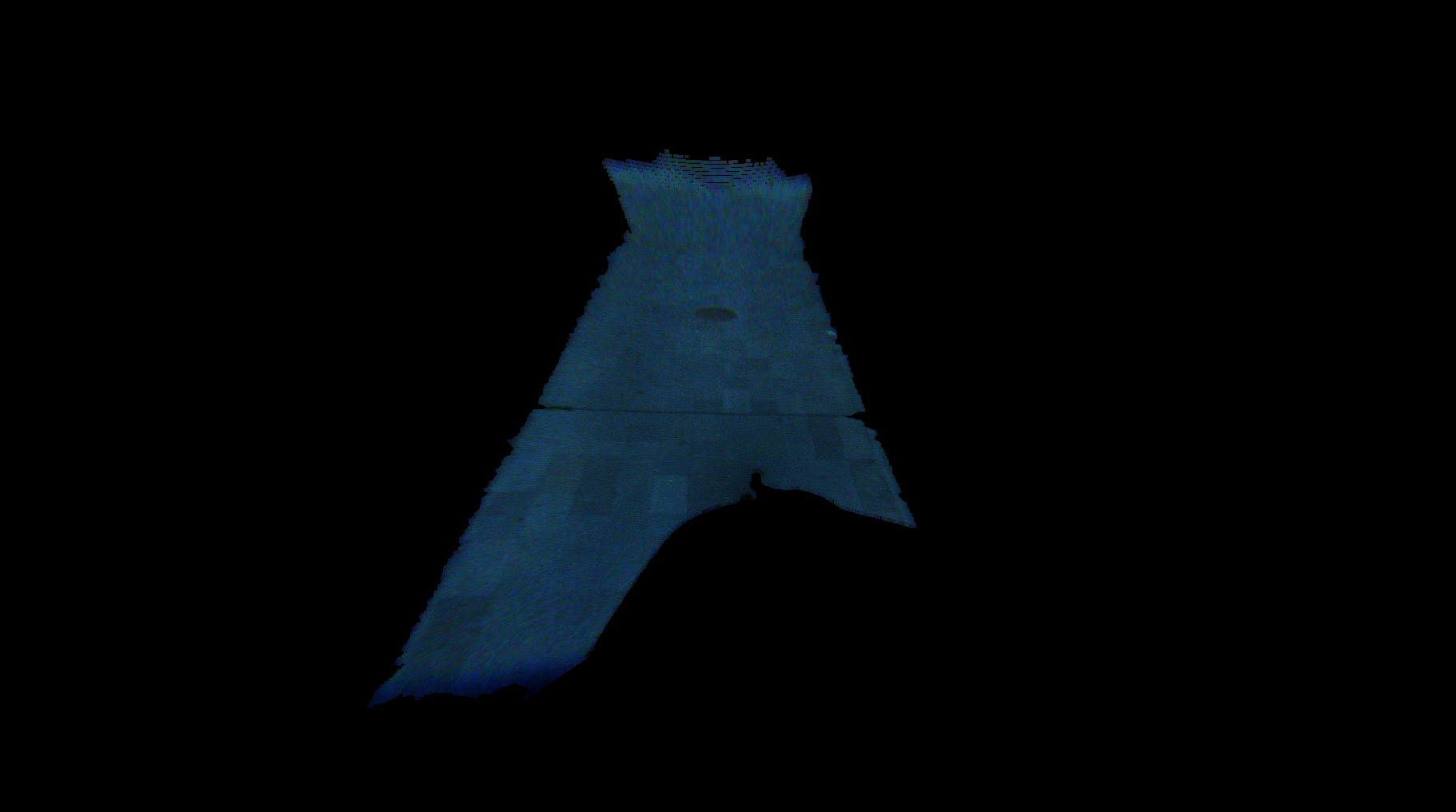}} \hfil    
    \subfloat{\includegraphics[width=0.24\textwidth]{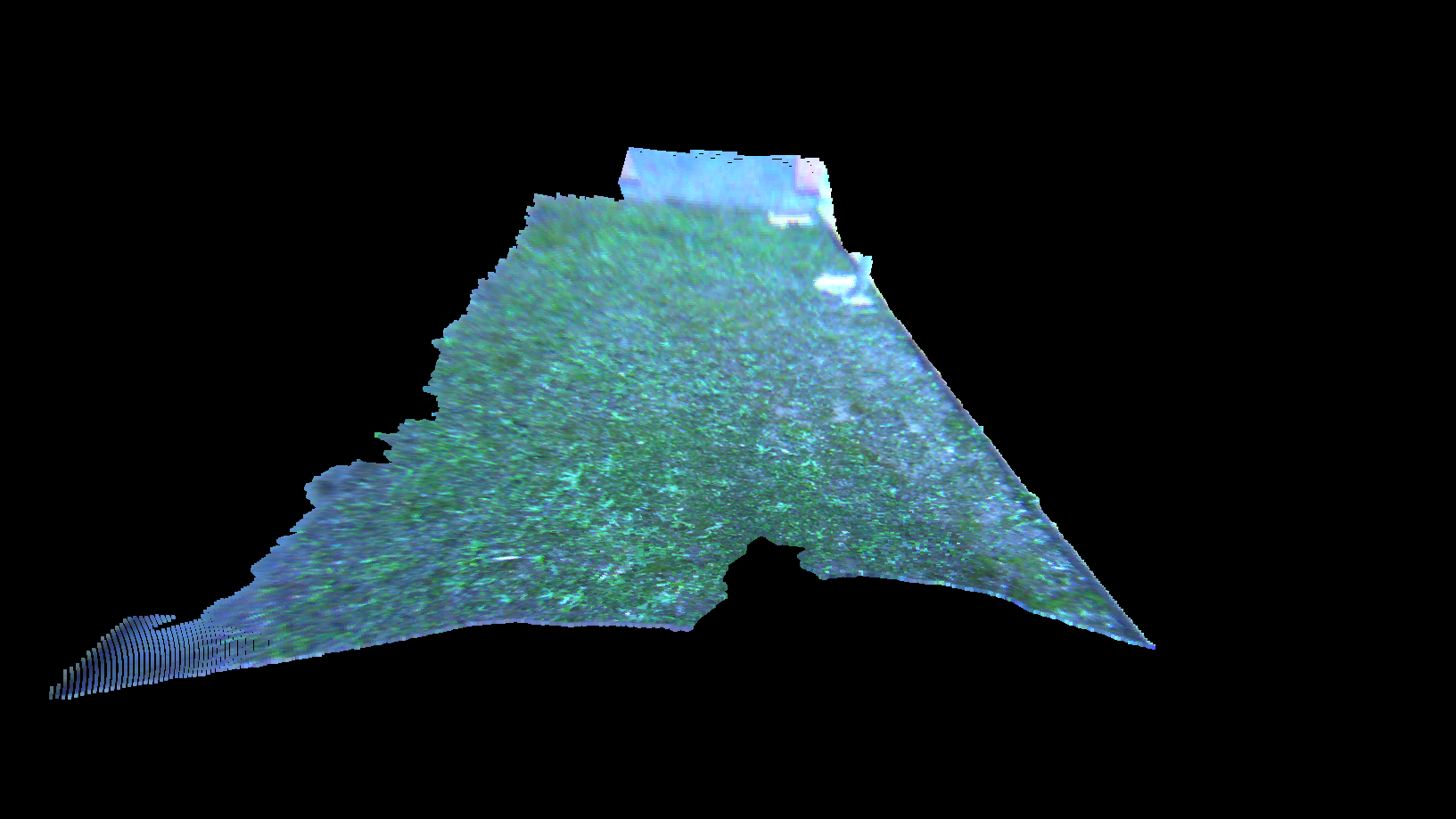}}

\caption{Images from the different experimental cases where the proposed method for floor extraction has been tested. In the first row the initial floor detection is represented; the second row shows the expanded floor segmentation and the third row is the 3D reconstruction of the obstacle-free space. Each column represents one case.}
\label{fig:experiments}
\end{figure*}

The floor extraction is tested for indoor and outdoor environments with different illumination conditions (see Fig. \ref{fig:experiments}). In order to obtain quantitative results in the floor extraction, in some frames of the sequences the floor has been manually labelled. In the indoor environments, there are four cases: {\it Case 1} and {\it Case 2} present environments with natural light of different intensity; {\it Case 3} presents a higher influence of artificial light; and {\it Case 4} shows lower illumination conditions. In the outdoor environments two cases are tested: on {\it Case 1} floor and walls have similar colours; and {\it Case 2} shows an irregular terrain. Since the depth sensor of the RGB-D camera is sensitive to direct sun light, outdoor experiments have been conducted on cloudy days. Fig. \ref{fig:in-c1} shows a frame of the Case 1 of the indoor environments, while Fig. \ref{fig:experiments} shows in each column frames of {\it Case 2} and {\it Case 3} of indoor environments and the outdoor environments cases.

\begin{figure*}[!t]
\centering   
	\subfloat{\includegraphics[width=0.24\textwidth]{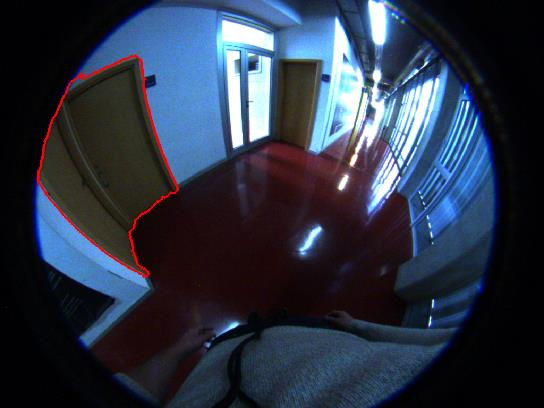}} \hfil
	\subfloat{\includegraphics[width=0.24\textwidth]{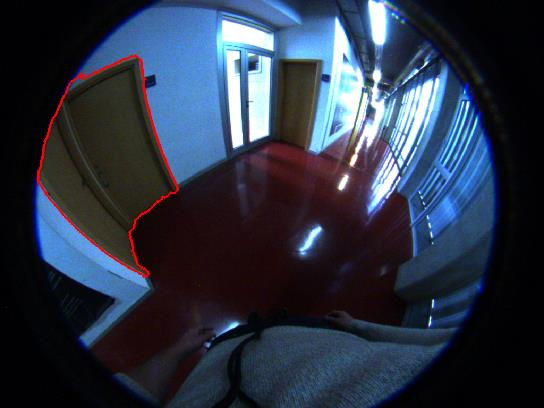}} \hfil
	\subfloat{\includegraphics[width=0.24\textwidth]{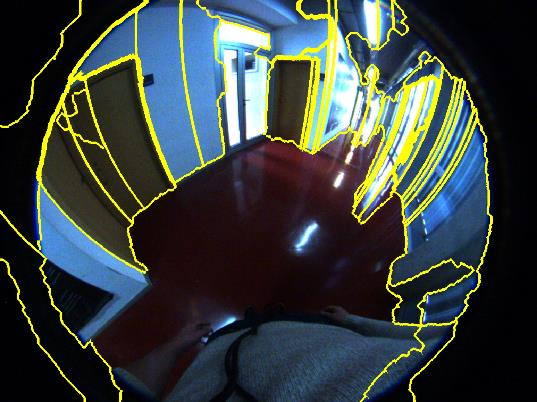}} \hfil
	\subfloat{\includegraphics[width=0.24\textwidth]{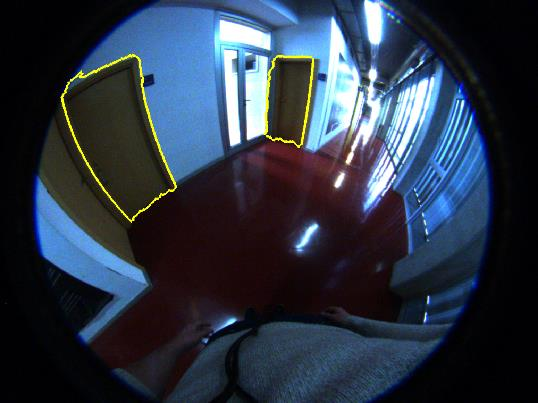}} \\ 
	\vspace{-2mm}		
	\subfloat{\includegraphics[width=0.24\textwidth]{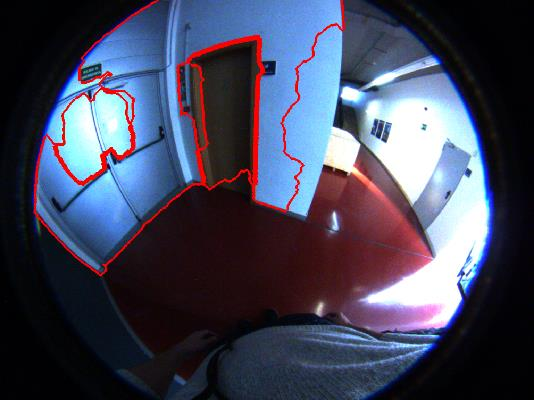}} \hfil
	\subfloat{\includegraphics[width=0.24\textwidth]{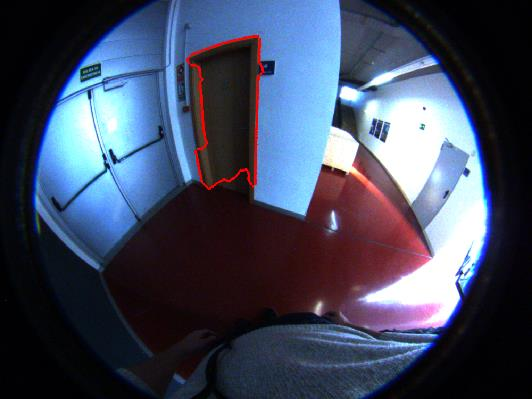}} \hfil
	\subfloat{\includegraphics[width=0.24\textwidth]{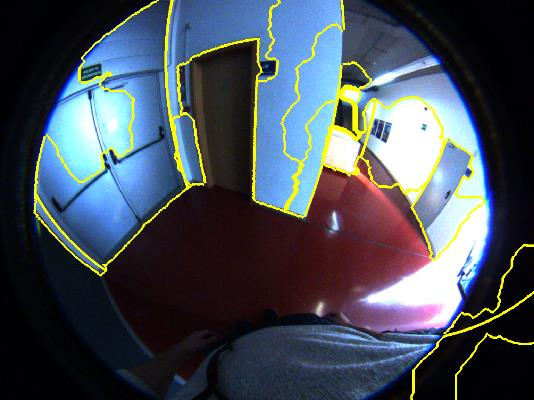}} \hfil
	\subfloat{\includegraphics[width=0.24\textwidth]{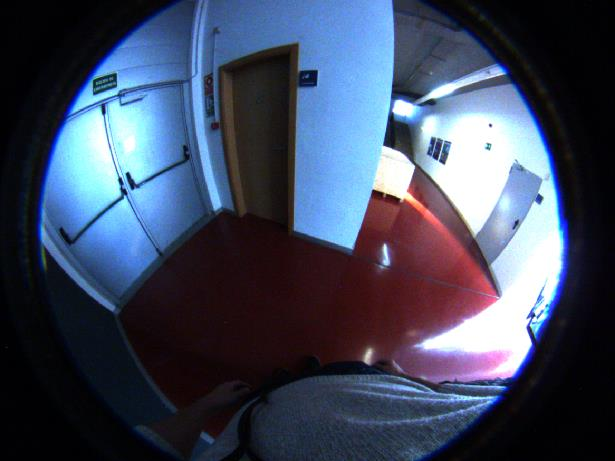}}   
	
\caption{Images from some experimental cases where the proposed method for door detection has been tested. The first and third columns represent the {\it Two-line} and {\it One-line} hypotheses respectively while the second and fourth columns represent the door detection for each hypotheses respectively. Each row represents one case.}
\label{fig:door_experiments}
\end{figure*}

The door detection algorithm is only tested for indoor environments (see Fig. \ref{fig:door_experiments}). As well as the floor extraction algorithm, different illumination conditions are tested to evaluate the performance of this approach. {\it Case 1} and {\it Case 2} present environments with mainly natural light while the rest of the cases present environments with mainly artificial light of different intensity.

\section{Results}
\label{sec:result}

In order to quantitatively evaluate the performance of the proposed algorithms, different parameters are defined for each experiment. 
For the door detection algorithm, it will be evaluated the percentage of doors of the scene which are included in the hypotheses (Hd), {\it Two-lines} or {\it One-line} hypotheses, and the number of detected doors (Dd) from those included in the hypotheses, as well as the running time of the algorithm for each hypothesis. This algorithm has been set to avoid all the false positives, so it is meaningless to measure its presicion or recall. Besides, since the algorithm is focused in door detection but not in door segmentation, the labelling is made qualitatively. It means that it is set the number of doors in the scene and the general area of them, but it is not taken into acount the exact shape of the door that the algorithm obtains.
On the other hand, to quatitatively evaluate the floor expansion, the floor has been manually labelled in some frames for each case as ground truth. The parameters defined for the evaluation of the floor extraction algorithm are: Precision (P) described as the correct expanded floor area over the total expanded floor area; Recall (R) defined as the correct expanded floor area over the floor area labelled as ground truth; area in square meters obtained with the RGB-D camera (A1); the total expanded area in square meters obtained with the combination of cameras (A2), and the ratio of floor expansion $AR = A2/A1$. 

\begin{table}[h]
	\centering
	\caption{Door detection results.}
	\label{tab:door_res}
	\begin{tabular}{c|ccc|ccc|}
	\cline{2-7}
          & \multicolumn{3}{c|}{{\it Two-line} hypothesis} & \multicolumn{3}{c|}{{\it One-line} hypothesis} \\
          					& time(s)& Hd (\%)& Dd (\%)& time(s)& Hd (\%)& Dd (\%)  \\ \hline
\multicolumn{1}{|c|}{Case1} &  1.6   &   50   &  100   &  4.24  &  100   &  33.33 \\
\multicolumn{1}{|c|}{Case2} &  2.23  &   50   &  100   &  2.25  &  100   &   50   \\
\multicolumn{1}{|c|}{Case3} &  3.02  &   75   &  100   &  5.9   &  100   &  66.67 \\
\multicolumn{1}{|c|}{Case4} &  0.86  &  33.33 &  100   &  16.31 &   75   &   50   \\
\multicolumn{1}{|c|}{Case5} &  2.65  &   75   & 33.33  &  5.82  &  100   &   25   \\ \hline
\multicolumn{1}{|c|}{$\Sigma$}  &  2.07  &  56.67 & 86.67  &  6.90  &   95   &   45   \\ \hline
	\end{tabular}
\end{table}

Table \ref{tab:door_res} presents the results of the proposed door detector. It can be observed that the {\it Two-line} hypothesis is faster than the {\it One-line} hypothesis, which seems reasonable since it is more restrictive and computes less hypotheses. Besides, the {\it Two-line} hypothesis detects more doors within the hypotheses it makes. However, in the hypotheses, only half of the doors of the scene are included against the {\it One-line} hypothesis algorithm that includes almost all of them. From these results, it can be concluded that the proposed algorithm can obtain almost half of the doors in the environment without false positives, which is really important in the guidance of visually impaired people. 

\begin{table}[!h]
  \centering
  \caption{Floor extraction results.}
  \label{tab:nav-result}
  \begin{tabular}{cc|ccccc|}
   \cline{3-7}
                                                &        &   P   &   R   &   A1 ($m^2$)  &   A2 ($m^2$)   &   AR   \\ \hline
   \multicolumn{1}{|c}{\multirow{4}{*}{Indoor}} &  Case1 & 0.996 & 0.916 & 2.951 & 29.828 & 10.108 \\
   \multicolumn{1}{|c}{}                        &  Case2 & 0.978 & 0.901 & 2.654 & 36.214 & 13.645 \\
   \multicolumn{1}{|c}{}                        &  Case3 & 0.992 & 0.952 & 3.131 & 26.211 & 8.371  \\
   \multicolumn{1}{|c}{}                        &  Case4 & 0.992 & 0.825 & 3.357 & 65.845 & 19.614 \\ \hline  
   \multicolumn{1}{|c}{\multirow{2}{*}{Outdoor}}&  Case1 & 0.994 & 0.984 & 3.149 & 60.649 & 19.259 \\
   \multicolumn{1}{|c}{}                        &  Case2 & 1.000 & 0.961 & 2.533 & 56.858 & 22.447 \\ \hline 
   \end{tabular}
\end{table}

From the floor extraction results presented in Table \ref{tab:nav-result} can be concluded that the hybrid system is able to extend the floor area obtained with a commercial RGB-D from 8 to 20 times in different environments, with great precision and recall. Besides, since the system is mounted on a wearable device, it works in both indoor and outdoor environments, improving other systems of the state of the art.

\section{Discussion}
\label{sec:discussion}

\begin{figure}[!t]
    \centering
    \subfloat[]{\includegraphics[width=0.24\textwidth]{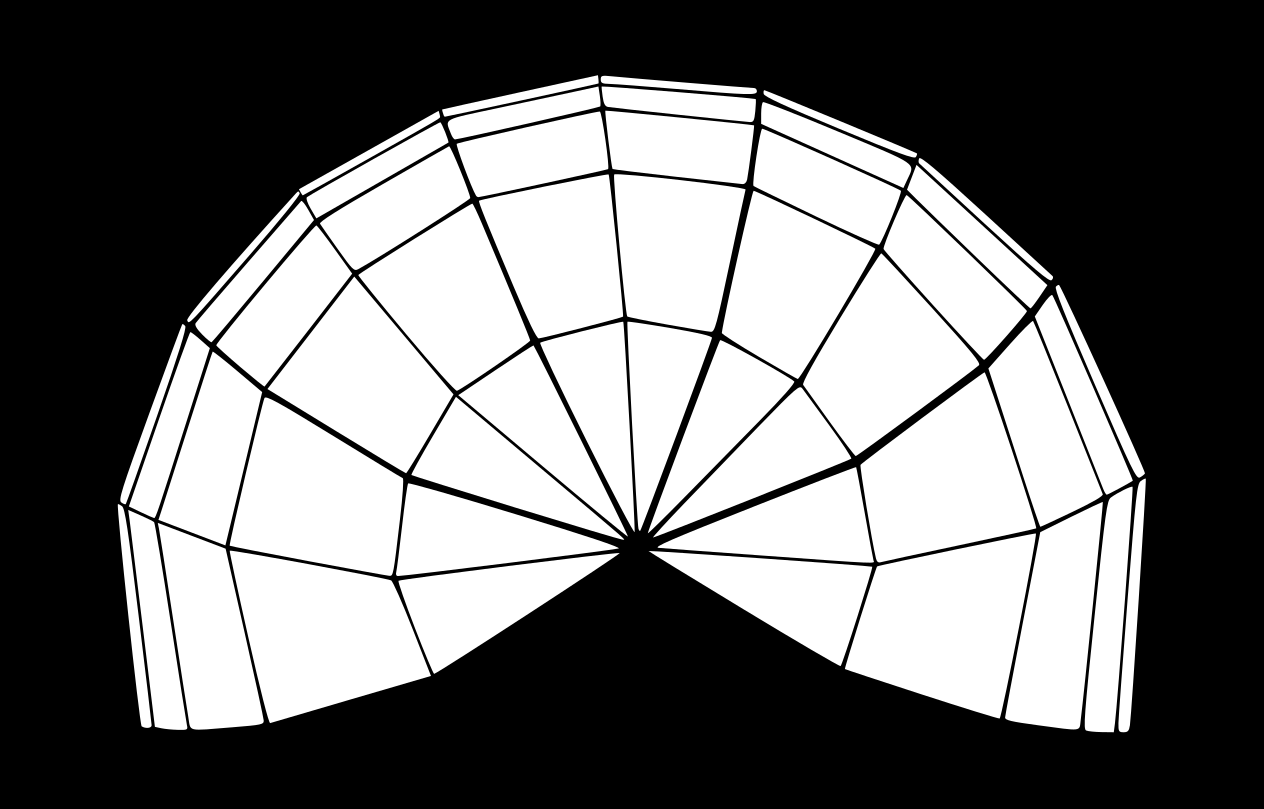} \label{fig:nav_sec}}
    \hfil
    \subfloat[]{\includegraphics[width=0.23\textwidth]{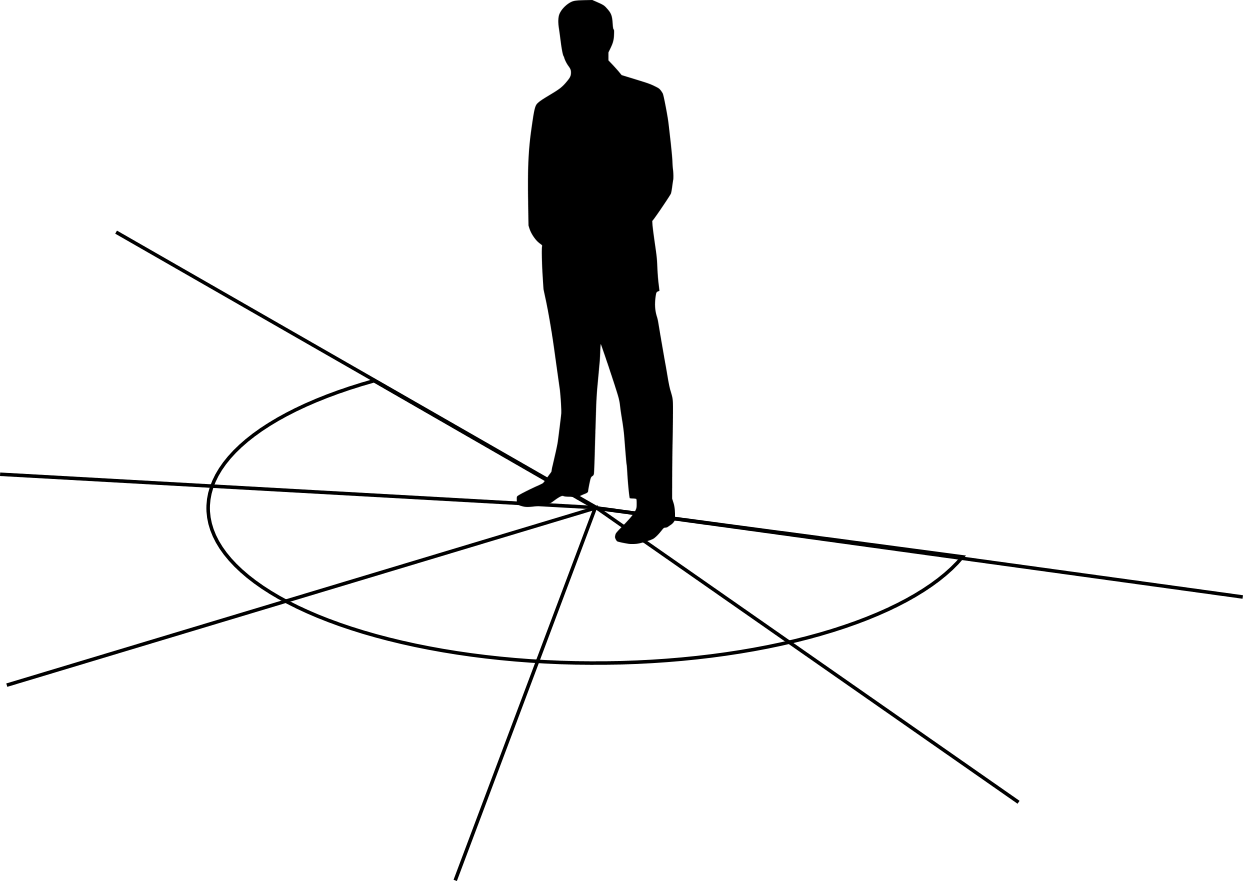} \label{fig:nav}}
    \caption{(a): Discretization in sectors of the obstacle-free space; (b): Navigation paths from the user point of view.}
    \label{fig:navigation}
\end{figure}

The results from the experiments show that the algorithms proposed in this work can achieve a floor segmentation and a door detection in different environments. In order to integrate the proposed method in a navigation system, a post-processing of the information obtained from the combination of cameras is proposed as well as a sketch of how to inform the user where is the free space.

First of all, the obstacle-free space should be obtained from the user's point of view with the information given by the algorithms proposed. For this purpose, the expanded floor is reprojected into an egocentric 3D space.
Since the projection model of the fish-eye camera has been computed in section \ref{sec:background}, casting the rays of the extended floor in the fish-eye image and computing the intersection with the floor plane computed in section \ref{subsec:floor-det} allows to obtain the free space in the environment. The obtained reprojected space is expressed in the camera reference system, where it is difficult to measure distances to obstacles. Changing the reference system to the floor plane, taking the user as the origin, allows to measure and express the distance from the user to the obstacles in a natural user centered reference (see Fig. \ref{fig:cam2floor}). With this configuration, it is proposed to discretize the movement directions in 5 sectors centered in the user, as seen in Fig. \ref{fig:nav}, which at the same time are split to have more robustness. The floor space is also discretized into 5 depth sectors (1 meter, 1-3 m, 3-7 m, 7-15 m, \textgreater 15 m), obtaining a discrete grid where is easy to define a path for the user (see Fig. \ref{fig:nav_sec}). By computing the distance to the closest obstacle for each sector, the free sectors can be determined and the information is provided to the user through the voice commands from Table \ref{table:audiocommands}.

\begin{table}[h]
    \centering
    \caption{Audio commands for the discrete space.}
    \label{table:audiocommands}
    \begin{tabular}{|c|l|c|l|}
        \hline
        Sector & Direction & Distance & Audio command \\ \hline
        1 &  Left & 1 m & Left 1 \\
        2 &  Front-left & 3 m & Front-left 3 \\
        3 &  Front & 7 m &  Front 7 \\
        4 &  Front-right & 15 m &  Front-right 15\\
        5 &  Right & \textgreater 15 m & Right max\\ \hline
    \end{tabular}
\end{table}

However, not all the extracted floor is accessible. Floor regions under objects, as tables or chairs, belong to the floor but the user cannot walk though them. These obstacles are eliminated from the navigable zones taking advantage of the 3D information. Taking the 3D points which stand out the floor plane and project them vertically, if the projection intersects with the extracted floor, that zone is removed from the navigable space.

\section{Conclusions}
\label{sec:conclusion}

This paper has presented a method to obtain obstacle-free zones from any environment for the assistance of visually impaired people. The method is composed by a hybrid camera system which takes advantage of the wide field of view of fish-eye cameras as well as depth information from RGB-D cameras. The combination of both sensors in a wearable device is able to obtain information around the user. A floor extraction algorithm that combines information obtained from both cameras has been presented, expanding the floor area extracted by a comercial RGB-D camera up to 20 times. Using this information of the surroundings, planning routes and trajectories instead of only avoiding obstacles can be done. Besides, a door detector is proposed, that uses the same system, allowing the navigation through different environments. The experiments show that both algorithms, floor extraction and door detector, perform with high precision and few to none false positives, creating safe routes for the navigation of visually impaired people or autonomous robots. These results encourage integration of these algorithms in navigation systems as the proposed for visually impaired people.

\section*{ACKNOWLEDGMENT}

Work supported by Projects RTI2018-096903-B-I00 and DPI2015-65962R (MCIU/AEI/FEDER, UE)

\bibliographystyle{unsrtnat}

\begin{thebibliography}{0}
\providecommand{\natexlab}[1]{#1}
\providecommand{\url}[1]{\texttt{#1}}
\expandafter\ifx\csname urlstyle\endcsname\relax
  \providecommand{\doi}[1]{doi: #1}\else
  \providecommand{\doi}{doi: \begingroup \urlstyle{rm}\Url}\fi

\end{thebibliography}


\begin{thebibliography}{99}

\bibitem{plikynas2020indoor}
{D. Plikynas, A. {\v{Z}}vironas, A. Budrionis and M. Gudauskis}. {\it Indoor Navigation Systems for Visually Impaired Persons: Mapping the Features of Existing Technologies to User Needs}.In {\em Sensors},{vol. 20},{num. 3},{p. 636}. {Multidisciplinary Digital Publishing Institute}, {2020}.

\bibitem{simoes2016blind}
{WCSS Sim{\~o}es and VF De Lucena}. {\it Blind user wearable audio assistance for indoor navigation based on visual markers and ultrasonic obstacle detection}. In {\em International Conference ICCE}, {p. 60--63}. {IEEE}, {2016}.

\bibitem{pereira2015blind}
{A. Pereira, N. Nunes, D. Vieira, N. Costa, H. Fernandes and J. Barroso}. {\it Blind guide: An ultrasound sensor-based body area network for guiding blind people}. In {\em Procedia Computer Science}, {vol. 67}, {p. 403--408}. {Elsevier}, {2015}.

\bibitem{dao2016indoor}
{TK Dao, TH Tran, TL Le, H. Vu, VT Nguyen, DK Mac, ND Do, and TT Pham}. {\it Indoor navigation assistance system for visually impaired people using multimodal technologies}. In {\em International Conference ICARCV}, {p. 1--6}. {IEEE}, {2016}.

\bibitem{islam2018indoor}
{MI Islam, MMH Raj, S. Nath, MF Rahman, S. Hossen and MH Imam}. {\it An Indoor Navigation System for Visually Impaired People Using a Path Finding Algorithm and a Wearable Cap}. In {\em International Conference I2CT}, {p. 1--6}. {IEEE}, {2018}.

\bibitem{guimaraes2013analysis}
{CSS Guimar{\~a}es, RVB Henriques and CE Pereira}. {\it Analysis and design of an embedded system to aid the navigation of the visually impaired}. In {\em Biosignals and Biorobotics Conference: Biosignals and Robotics for Better and Safer Living}, {p. 1--6}. {IEEE}, {2013}.

\bibitem{yang2018long}
{K. Yang, K. Wang, S. Lin, J. Bai, LM Bergasa and R, Arroyo}. {\it Long-range traversability awareness and low-lying obstacle negotiation with RealSense for the visually impaired}. In {\em International Conference ICISS}, {p. 137--141}. {2018}

\bibitem{han2015fuzzy}
{SB. Han, DH. Kim and JH. Kim}. {\it Fuzzy gaze control-based navigational assistance system for visually impaired people in a dynamic indoor environment}. In {\em International Conference FUZZ}, {p. 1--7}. {IEEE}, {2015}.

\bibitem{zhang2015slam}
{X. Zhang, B. Li, SL. Joseph, J. Xiao, Y. Sun, Y. Tian, JP. Mu{\~n}oz and C. Yi}. {\it A slam based semantic indoor navigation system for visually impaired users}. In {\em International Conference SMC}, {p. 1458--1463}. {IEEE}, {2015}.

\bibitem{bai2017smart}
{J. Bai, S. Lian, Z. Liu, K. Wang and D. Liu}. {\it Smart guiding glasses for visually impaired people in indoor environment}. In {\em Transactions on Consumer Electronics}, {vol. 63}, {num. 3}, {p. 258--266}. {IEEE}, {2017}.

\bibitem{chen2017ccny}
{Q. Chen, M. Khan, C. Tsangouri, C. Yang, B. Li, J. Xiao and Z. Zhu}. {\it CCNY smart cane}. In {\em International Conference CYBER}, {p. 1246--1251}. {IEEE}, {2017}.

\bibitem{lee2016rgb}
{YH. Lee and G. Medioni}. {\it RGB-D camera based wearable navigation system for the visually impaired}. In {\em Computer Vision and Image Understanding}, {vol. 149}, {p. 3--20}. {Elsevier}, {2016}.

\bibitem{zeineldin2016fast}
{RA Zeineldin and NA El-Fishawy}. {\it Fast and accurate ground plane detection for the visually impaired from 3D organized point clouds}. In {\em SAI Computing Conference}, {p. 373--379}. {IEEE}, {2016}.

\bibitem{peasley2013real}
{B. Peasley and S. Birchfield}. {\it Real-time obstacle detection and avoidance in the presence of specular surfaces using an active 3D sensor}. In {\em Workshop on Robot Vision}, {p. 197--202}. {IEEE}, {2013}.

\bibitem{wang2014segment}
{Z. Wang, H. Liu, X. Wang and Y. Qian}. {\it Segment and label indoor scene based on RGB-D for the visually impaired}. In {\em International Conference on Multimedia Modeling}, {p. 449--460}. {Springer}, {2014}.

\bibitem{yang2016expanding}
{K. Yang, K. Wang, W. Hu and J. Bai}. {\it Expanding the detection of traversable area with RealSense for the visually impaired}. In {\em Sensors}, {vol. 16}, {num. 11}, {p. 1954}. {Multidisciplinary Digital Publishing Institute}, {2016}.

\bibitem{pham2016real}
{HH Pham, TL Le and N. Vuillerme}. {\it Real-time obstacle detection system in indoor environment for the visually impaired using microsoft kinect sensor}. In {\em Journal of Sensors}, {vol. 2016}. {Hindawi}, {2016}.

\bibitem{aladren2014navigation}
{A. Aladren, G. L{\'o}pez-Nicol{\'a}s, L. Puig and J.J. Guerrero}. {\it Navigation assistance for the visually impaired using RGB-D sensor with range expansion}. In {\em Systems Journal}, {vol. 10}, {num. 3}, {p. 922--932}. {IEEE}, {2016}.

\bibitem{perez2016peripheral}
{A. Perez-Yus, G. Lopez-Nicolas and J.J. Guerrero}. {\it Peripheral expansion of depth information via layout estimation with fisheye camera}. In {\em ECCV}, {p. 396--412}. {Springer}, {2016}.

\bibitem{perez2016novel}
{A. Perez-Yus, G. Lopez-Nicolas and J.J. Guerrero}. {\it A novel hybrid camera system with depth and fisheye cameras}. In {\em International Conference ICPR}, {p. 2789--2794}. {IEEE}, {2016}.

\bibitem{scaramuzza2006toolbox}
{D. Scaramuzza, A. Martinelli and R. Siegwart}. {\it A toolbox for easily calibrating omnidirectional cameras}.In {\em International Conference IROS},{p. 5695--5701}. {IEEE}, {2006}.

\bibitem{perez2017extrinsic}
{A. Perez-Yus, E. Fernandez-Moral, G. Lopez-Nicolas, J.J. Guerrero and P. Rives}. {\it Extrinsic calibration of multiple RGB-D cameras from line observations}. In {\em Robotics and Automation Letters}, {vol. 3}, {num. 1}, {p. 273--280}. {IEEE}, {2018}.

\bibitem{bermudez2015automatic}
{J. Bermudez-Cameo, G. Lopez-Nicolas and J.J. Guerrero}. {\it Automatic line extraction in uncalibrated omnidirectional cameras with revolution symmetry}. In {\em IJCV}, {vol. 114}, {num. 1}, {p. 16--37}. {Springer}, {2015}.

\bibitem{perez2019scaled}
{A. Perez-Yus, G. Lopez-Nicolas and J.J. Guerrero}. {\it Scaled layout recovery with wide field of view RGB-D}. In {\em Image and Vision Computing}, {vol. 87}, {p. 76--96}. {Elsevier}, {2019}.

\bibitem{van2012seeds}
{M. Van den Bergh, X. Boix, G. Roig,B. de Capitani and L. Van Gool}. {\it Seeds: Superpixels extracted via energy-driven sampling}. In {\em ECCV}, {p. 13--26}. {Springer}, {2012}.

\bibitem{murillo2008visual}
{AC. Murillo, J. Ko{\v{s}}eck{\'a}, J.J. Guerrero and C. Sag{\"u}{\'e}s}. {\it Visual door detection integrating appearance and shape cues}. In {\em Robotics and Autonomous Systems}, {vol. 56}, {num. 6}, {p. 512--521}. {Elsevier}, {2008}.

\end{thebibliography}

\end{document}